%% file: main_database.tex
\documentclass[12pt]{article}
\usepackage{epsfig}
\usepackage{amssymb}

\usepackage{setspace} % to adjust spacing
\usepackage{amsmath}
\usepackage[utf8]{inputenc}
\usepackage{booktabs} % To thicken table lines
\usepackage{subfig}
\usepackage{float}
\usepackage{multirow}
\usepackage{placeins}
\usepackage{hyperref}
\usepackage{geometry}
\usepackage{authblk}

\newcommand\blfootnote[1]{%
  \begingroup
  \renewcommand\thefootnote{}\footnote{#1}%
  \addtocounter{footnote}{-1}%
  \endgroup
}

\usepackage{url}
\usepackage{natbib}
\usepackage{tikz}
\usepackage{tkz-graph}  
\usetikzlibrary{shapes.geometric}
\usetikzlibrary{matrix, positioning, fit}

\newcommand\numberthis{\addtocounter{equation}{1}\tag{\theequation}}

\begin{document}

    \begin{titlepage}
        \title{{DARE}: A large-scale handwritten {da}te {re}cognition system\blfootnote{Acknowledgements: Torben gratefully acknowledges financial support from the Independent Research Fund Denmark, grant 8106-00003B. Emil gratefully acknowledges financial support from the European Research Council (Starting Grant Reference 851725). The DARE database is available at \url{https://www.kaggle.com/datasets/sdusimonwittrock/dare-database}. Our code is available upon request.}}
%% use optional labels to link authors explicitly to addresses:
%% \author[label1,label2]{}
%% \affiliation[label1]{organization={},
%%             addressline={},
%%             city={},
%%             postcode={},
%%             state={},
%%             country={}}
%%
%% \affiliation[label2]{organization={},
%%             addressline={},
%%             city={},
%%             postcode={},
%%             state={},
%%             country={}}
        
        \author[*]{Christian M. Dahl}
        \author[*]{Torben S. D. Johansen}
        \author[**]{Emil N. Sørensen}
        \author[*]{Christian E. Westermann}
        \author[*]{Simon F.  Wittrock}

        \affil[*]{Department of Economics, University of Southern Denmark}%Department and Organization
        \affil[**]{School of Economics, University of Bristol}
        \maketitle
        
        \begin{abstract}
        \input{tex/0_abstract.tex}
        \end{abstract}
        
        \section*{Keywords}
        Handwritten date recognition; EfficientNetV2; Multi-head classification; Transfer learning; Date database. 
        
    \end{titlepage}
    
\doublespacing

\input{./tex/1_introduction.tex}

\input{./tex/2_database.tex}

\input{./tex/3_results.tex}

\input{./tex/4_discussion_and_conclusion}
%\input{./tex/old-result-tables}
\newpage % added 
%% The Appendices part is started with the command \appendix;
%% appendix sections are then done as normal sections
\onehalfspacing
\bibliographystyle{plainnat}
\bibliography{bib}

\appendix

\newpage % added 
\doublespacing

\input{./tex/7_appendix.tex}

% Bibliography

%\newpage 
%\bibliography{bib}

\end{document}

%% file: tex/0_abstract.tex
\noindent
Handwritten text recognition for historical documents is an important task but it remains difficult due to a lack of sufficient training data in combination with a large variability of writing styles and degradation of historical documents.  
While recurrent neural network architectures are commonly used for handwritten text recognition, they are often computationally expensive to train and the benefit of recurrence drastically differs by task.
For these reasons, it is important to consider non-recurrent architectures.
%In this work, we show one such setting, 
In the context of handwritten date recognition,
we propose an architecture based on the EfficientNetV2 class of models that is fast to train, robust to parameter choices, and accurately transcribes handwritten dates from a number of sources.
For training, we introduce a database containing almost 10 million tokens, originating from more than 2.2 million handwritten dates which are segmented from different historical documents. 
As dates are some of the most common information on historical documents, and with historical archives containing millions of such documents, the efficient and automatic transcription of dates % is scalable and 
has the potential to lead to significant cost-savings over manual transcription. 
We show that training on handwritten text with high variability in writing styles result in robust models for general handwritten text recognition and that transfer learning from the DARE system increases transcription accuracy substantially, allowing one to obtain high accuracy even when using a relatively small training sample.
%Finally, based on the sparsely annotated 1916 Danish Census, we illustrate how to use the DARE system for linking and creating a unique historical database comprising more than 3.7 million entries with information on name, birth date, residence, civil status, income, and wealth for every individual living in Denmark (outside Copenhagen) in 1916. 

%In this paper, we present the DARE system, a handwritten date recognition system achieving human-level (and above) transcription accuracy on many historical documents. 
%The DARE system is trained on almost 10 million tokens, originating from more than 2.2 million handwritten dates which we segmented from different historical documents.
% We show that high variation text fosters more robust models for general handwritten text recognition. 
%Furthermore, we illustrate two challenges that the DARE system can help overcome: 
%First, when training new date recognition models, one has to label a large amount of data to obtain a useful training sample. 
%We show that transfer learning from the DARE system increases transcription accuracy drastically, allowing one to obtain high accuracy even when a relatively small training sample is available.
%Second, we illustrate how to use the DARE system in a zero-shot fashion in combination with transfer learning, to link more than 140 thousand observations from the 1916 Danish census. This facilitates training of models for transcribing the entire 1916 census of 3.7 million observations.
%or This in turn allows us to train models to transcribe the entire 1916 census of 3.7 million observations.}

%% file: tex/1_introduction.tex
%More efficient and accurate transcription methods  have enormous potential to reduce costs and increase quality of downstream research relying on transcribed dates. Dates are present on a wide variety of historical documents. Transcribing these are required to answer a large number of research questions, such as when studying historical mortality patterns, where date of death is directly needed (see, e.g., \citet{Viboud2013, Dahal2018}).
%Further, researchers are often interested in following individual across, e.g., censuses, and historical linking methods often rely on birth dates of individuals, either directly or obtained by subtracting their age from the year of the census (see, e.g., \citet{USCensus2018}).
%While manual transcription is sometimes feasible, it is both expensive and slow \citep{Toselli2010, Romero2017}, and the alternative, automated methods such as Transkribus, are often not able to accurately transcribe dates from handwritten historical documents with a tabular form \citep{Muehlberger2019, dahl2021applications}. 
\section{Introduction}
Handwritten text recognition (HTR) is an important step towards the transcription and preservation of historical archives but it remains a challenging task. 
One obstacle is that deep learning models often require large-scale annotated datasets to perform well, which is particularly a concern for HTR on historical documents due to the high variation in writing styles and that historical documents often suffer from bleed-through, fading of text, and general degradation. 
One step towards solving this is to introduce more and larger real-world datasets that can increase the robustness of HTR models. 
Even though dates are some of the most important and frequently represented information on historical documents, little research have been invested into the specific recognition of handwritten dates from historical documents. 
To solve the challenge of handwritten date transcription from historical documents, we introduce the DARE database and system, which together facilitate the transcription of dates even from difficult-to-read historical documents.\footnote{While general-purpose out-of-the-box HTR such as Transkribus has improved significantly in recent years, transcribing dates from historical documents proves challenging due to issues including document degradation, other handwriting overlapping dates in segmented images, and variation in handwriting styles.} 
The DARE database is the largest available database of handwritten dates and consists of segmented dates from a number of different historical documents, meaning that the database contains many types of documents from different time periods written by a large number of authors and suffering from varying amounts of degradation.
This results in significant variation between the handwritten dates, see, for example, Figure \ref{fig: database-image-examples-33}, which enables us to train robust models for transcription, which in turn perform well for transfer learning.
In total, the DARE database is derived from 3,145,922 images, originating from six different data sources.
While a large number of the segmented images are empty, we still obtain more than 2.2 million manually transcribed dates, totalling almost 10 million individual tokens.\footnote{We still include the empty images when training our models, as we want models that are robust in the sense of being able to inform when a date is \textit{not} present on an image.}  %2,221,954 9,932,010 tokens
%We segment the regions of interest (the dates) from these documents to construct the DARE database.

We use this large database of handwritten dates to train neural networks to transcribe dates: The DARE system.
%We train neural networks on different subsets of the DARE database. %and for slightly different purposes which depends on how much of the dates require transcription, i.e., sometimes it is not necessary to transcribe the year.
These networks achieve date transcription accuracies of between 92\% to 99\%, with the exception of one particularly difficult dataset.\footnote{This is primarily due to errors in the manual labels and the segmentation. However, we still believe that this dataset is helpful to improve the overall performance of our neural networks.}
To put this performance into context, we compare the performance on one of these datasets to previous work and demonstrate significant improvements over manual labelling.
%On this dataset, we demonstrate significant improvements over both manual and automatic labelling, i.e. %we achieve higher transcription performance relative to the manual transcription performance discussed in \citet{goodfellow2013multi} and \citet{dahl2021applications} and higher than other automatic transcription procedures on \textbf{DC-1} presented in \citet{dahl2021applications}.
%On this dataset, we demonstrate significant improvements over manual labelling.%, where we achieve higher transcription performance relative to the manual transcription performance in \citet{dahl2021applications}. We also achieve significantly higher transcription accuracy than the automatic transcription network suggested in \citet{dahl2021applications}.
While it is reassuring that the DARE system achieves high levels of transcription accuracies on the test sets, our ultimate objective is to create a system that improves automated transcription of dates from new data sources.
To illustrate the usefulness of the DARE system for such tasks, we first use the DARE system for transfer learning, showing that it significantly improves transcription performance over networks trained from scratch as well as networks transfer learned on other data sources.
Using the DARE system, we are able to achieve high levels of transcription accuracy even when using only a small training sample.

Second, we use Danish census data from 1916 to further illustrate the usefulness of the DARE system for transfer learning and for zero-shot transcription.
A subset (less than 5\%) of the 3.7 million entries recorded in the 1916 census, for which we have image files, are labelled manually. However, the labels come with no link back to the source images, so the labels cannot directly be used as training data.
We show that by transcribing images in a zero-shot fashion using the DARE system and subsequently linking the images to the labelled data, we can gradually create a larger training set to which we can eventually apply transfer learning and obtain a more accurate model. This procedure is repeated until a sufficient degree of accuracy is obtained. % For the linking, we propose a strategy that simultaneously attempts to remove false matches while also increasing the match rate, by iteratively transcribing dates, matching entries to the labelled data, and increasing the training sample.% and repeating this cycle multiple times.

This procedure allows us to match nearly all of the labelled entries to the source images, resulting in a large training sample that makes it possible to transcribe the census with a date recognition accuracy of around 95\%. The resulting database comprising more than 3.7 million rows contains unique information on name, birth date, residence,  civil status, income, and wealth for every single individual living in Denmark (excluding Copenhagen residents) in 1916.

The rest of the paper proceeds as follows: %Section~\ref{sec: Constructing the DARE Database} describes the database and the data acquisition procedure.
We start by surveying some of the most relevant literature within HTR.
Next, in Section \ref{sec: DARE System} we describe the database, network architecture, and technical details of the training pipeline.
Section \ref{sec: transcript performance}
shows the within-distribution performance of the DARE system, by studying the performance of the system on the test sets of the DARE database. 
Section~\ref{sec: tl} describes the pipeline for transfer learning and presents the associated results. 
Section~\ref{sec: linking} shows the process of linking images to transcriptions.
Section~\ref{sec: discussion-and-conclusion} concludes. 
%We provide additional details of the different data sources used to create the database in \ref{sec: Details on Data Sources}.

%% file: tex/2_database.tex
\section{Related work}
Performance of state-of-the-art deep learning HTR models have improved considerably within recent years.
From an initial focus on hidden Markov models (HMM) \citep{Toselli2004, Bertolami2008}, more recently, focus has shifted towards deep learning models such as CNNs \citep{Yousef2020}, hybrid CNN-RNNs \citep{Dutta2018}, or other types of RNNs, including LSTM models for HTR challenges, such as bidirectional LSTMs (BLSTM) \citep{Graves2008b, Curtis2018}, multidimensional LSTMs (MDLSTM) \citep{Graves2008}, MDLSTM-RNNs \citep{Puigcerver2017}, or deformable convolutional-recurrent networks \citep{Cascianelli2022}.
The introduction of the connectionist temporal classification (CTC) loss \citep{Graves2006} played a significant role and provided a significant advantage for RNNs over the previous HMMs \citep{Curtis2018}. 
However, one of the challenges with many of the different variants of the recurrent models is that they are computationally slow due to the lack of parallelization during training \citep{Kang2022}.

Another limiting factor for deep learning models, especially for HTR models where there is a huge variability of writing styles, is the need for large amounts of training data. 
To overcome this, it has been suggested to perform better data augmentation in order to increase the variability of the training data \citep{Puigcerver2017} and to enhance existing pre-processing procedures to mitigate some of the concerns regarding the degradation of historical documents, e.g., through improving upon the grayscale algorithm as proposed by \citet{Bouillon2019}.
Another possibility to overcome the lack of training data and reduce the cost of manual labelling is to use GANs \citep{Goodfellow2014GAN} as a semi-supervised tool to create synthetic handwritten text  \citep{Fogel2020}.

Inspired by the more recent interest in sequence-to-sequence models for HTR \citep{Aberdam2021, Geetha2021}, Transformers \citep{Vaswani2017} have also received an increased interest. 
Transformer-based models for HTR challenges are presented in both \citet{Li2021} and \citet{Kang2022} from where it seems that in order to obtain results similar to those acquired by other architectures, these currently require pre-training on large synthetic datasets followed by fine-tuning on real annotated data.

% Similar to Transformer-based models, this work also depart from much of the HTR...
This paper departs from much of the HTR literature by proposing a non-recurrent architecture.
Our focus is on a relatively light-weight architecture that is fast to train and performs well in the absence of large amounts of labelled data.
To achieve this, we draw inspiration from the literature on image classification and adopt many of the optimization techniques used there, e.g., RandAugment \citep{cubuk2020randaugment}, label smoothing \citep{szegedy2016rethinking},  %dropout \citep{srivastava2014dropout}, 
and stochastic depth \citep{huang2016deep}.
We use an EfficientNetV2 \citep{tan2021efficientnetv2} as the backbone of the architecture and a multi-head classification approach inspired by the house number recognition model proposed in \citet{goodfellow2013multi}. Our work shows that the non-recurrent architecture we propose is not a limitation in the setting of date recognition, which we believe is due to the limited additional information on the next token in a date sequence by knowing the former token.\footnote{For example, knowing that it is the 14th provides limited information about which month it is. In certain cases the reverse is true: For example, knowing that it is the 31st excludes a large number of months.} Instead it leads to faster training and inference due to additional available parallelization.

%% file: tex/3_results.tex
\section{The DARE system}
\label{sec: DARE System}
The DARE system we are proposing consists of several unique neural networks. Each of the networks are trained on specific subsets of the DARE database and each network has a unique task in relation to transcribing the dates.
%These models are useful for two purposes:
%First, they can be used to directly transcribe dates.
%Second, they can be used for transfer learning to improve transcription performance on new tasks, when some training data is available.
In this section, we present the DARE database, the neural networks, including their architecture and how they are trained.
%In Section \ref{tab: transcription-performance} we provide the transcription result of our model. 
%We then provide results for each network across a number of settings.
%In Section~\ref{sec: tl}, we illustrate how the DARE system increases transcription accuracy on new tasks, through the use of transfer learning. In Section~\ref{sec: linking}, we show how the DARE system can facilitate linking by transcribing dates from new images directly in a zero-shot fashion and then use transfer learning from the DARE system to provide accurate transcription which in turn increases the linking performance.

\subsection{Database}
\begin{figure}
    \centering
    \caption{Date Style Variance: August 28, 33}
    \label{fig: database-image-examples-33}
    \subfloat[DC-1]{
        \includegraphics[height=0.083\textwidth]{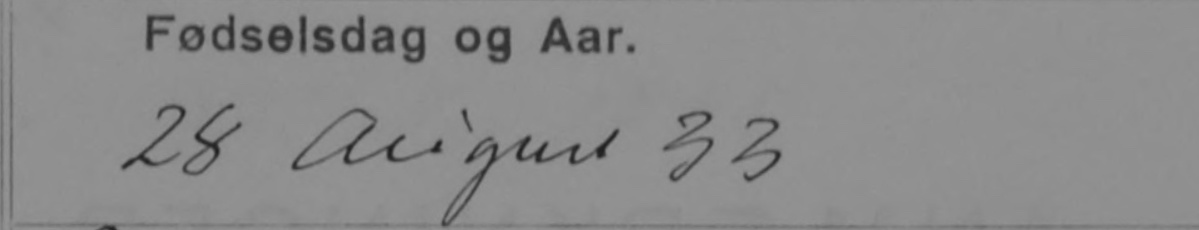}
        \label{fig:examplesDC-1-33}
    }
    \qquad
    \subfloat[PR-1]{
        \includegraphics[height=0.083\textwidth]{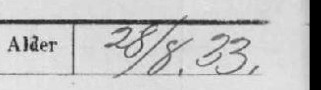}
        \label{fig:examplesPR-1-33}
    }
    \qquad
    \subfloat[SWE-BD]{
        \includegraphics[height=0.083\textwidth]{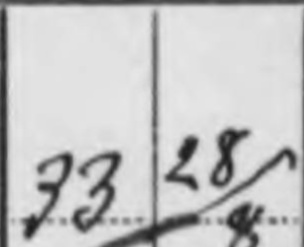}
        \label{fig:examplesSWE-BD-33}
    }
    \qquad
    \begin{minipage}{1\linewidth}
        \vspace{1ex}
        \scriptsize{
        \textit{Notes:}
        The figure shows a collection of examples from the DARE database with identical labels (the ground truth date for all three examples is: 28-08-33). 
        Panel~\ref{fig:examplesDC-1-33} shows an example where the day is written leftmost, the month in the middle as text, and the year rightmost.
        Panel~\ref{fig:examplesPR-1-33} shows an example where the day is written leftmost, the month in the middle as a number, and the year rightmost.
        Panel~\ref{fig:examplesSWE-BD-33} shows an example where the year is written leftmost, the day in the middle, and the month rightmost as a number.
        % Panel~\ref{fig:examplesSWE-1-33} differs in the sense that while the order of the components remains the same, the month is now written as a number; and Panel~\ref{fig:examplesSWE-BD-33} differs in the sense that now the year is written leftmost with the month still being written as a number.
        }
    \end{minipage}
\end{figure}

The DARE database consists of 3,145,922 images, of which 2,221,954 contain handwritten dates.
The handwritten dates are typically written on a single line.
The source images vary in size, with widths up to 1400 pixels and heights up to 356 pixels.
Figure~\ref{fig: database-image-examples-33} illustrates the large variance in style. We briefly describe our data acquisition strategy in \ref{sec: Details on Data Sources}. %provides more descriptions and examples of the datasets.
The visualization in \ref{sec: Details on Data Sources} together with Figure~\ref{fig: database-image-examples-33} illustrates that both within and between datasets, the style -- in terms of handwriting, structure, and content -- varies significantly.

Table \ref{tab:database} shows an overview of the different datasets, including what information is available (i.e., how complete the date is), which source the dataset originates from, and how many labelled images are in each dataset (split into a train and test sample). We split the database into a train and test set for each dataset separately.
The test set is randomly selected and consists of approximately 5\% of each dataset. 
The split enables us to test the performance on each dataset internally, which is important as the difficulty (in part due to image, segmentation, and label quality) differ between the datasets. There is also significant variation between the different handwritten dates across the sources, time-period and writing styles, of which an example is provided in Figure \ref{fig: database-image-examples-33}.

For ease of readability, we use an abbreviated name for each dataset (also visible in the table).
The general format of the date sequences across data sources is DD-M-YYYY whenever the year, including the century, is present on the images.\footnote{However, on the images this is not always the order, and, e.g., the year might be the first part of the date, when read left to right.} For many of the datasets part of (or the entire) year is not present: When only the last two digits of the year are available, we write DD-M-YY, and when the entire year is missing, we write DD-M.

\begin{table}[!ht]
    \centering
    \footnotesize{
    \caption{DARE Database: Datasets Details}
    \label{tab:database}
    \resizebox{1\linewidth}{!}{
        \begin{tabular}{lllrr}
            \toprule
            \textbf{Dataset} & \textbf{Data Source} & \textbf{Sequence} & \textbf{Train Observations} & \textbf{Test Observations} \\
            \midrule
            \textbf{DC-1} & Danish death certificates &  DD-M-YYYY & 11,627  & 1,000 \\
            \textbf{DC-2} & Danish death certificates & DD-M-YYYY & 155,439 & 8,338 \\
            \textbf{PR-1} & Danish police records & DD-M-YY & 1,006,199 & 53,488 \\
            \textbf{PR-2} & Danish police records & DD-M-YY & 326,478 & 17,103 \\
            \textbf{SWE-BD} & Swedish causes of death register & DD-M-YY & 597,756  & 31,389 \\
            \textbf{FR} & Danish funeral records & DD-M & 231,440 & 12,293 \\
            \textbf{NHVD} & Danish nurse home visiting documents & DD-M & 106,162 & 11,874 \\
            \textbf{SWE-DD} & Swedish causes of death register & DD-M & 547,813 & 28,803 \\
            \bottomrule
        \end{tabular}}
    }
    \begin{minipage}{1\linewidth}
        \vspace{1ex}
        \scriptsize{
            \textit{Notes:}
            The table shows an overview of the datasets of the DARE database, with the third column showing what part of the date sequence is available for the different datasets.
            In addition, we provide an overview of the training and evaluation dataset sizes, which range from 11,627 to 1,006,199 training observations and 1000 to 53,488 test observations.
            The datasets are ordered according to the length of the date sequence.
            Note that, due to data confidentiality, the Danish nurse home visiting documents are not publicly available.
            }
    \end{minipage}
\end{table}

\subsection{Network architecture}
Each neural network uses an EfficientNetV2-S as its feature extractor \citep{tan2021efficientnetv2}.
This choice is motivated by the unprecedented trade-off between accuracy and required computation achieved by the family of EfficientNetV2 models, both compared to other convolutional neural networks and vision transformers and both in terms of training and inference speed. While our models are largely identical, the networks differ in terms of their classification heads.
As shown in Table~\ref{tab:database}, the exact date information available differs across the different datasets. Occasionally, only the day and month is available, and sometimes we have access to day, month, and part of the year, while other times the full date sequence is present on the image. For this reason, we construct networks with three different architectures: An architecture that only predicts the day and month, an architecture that predicts the day, month, and the last two digits of the year, and an architecture that predicts the day, month, and entire (four digits) year.

To construct classification heads able to predict dates, we draw inspiration from \citet{goodfellow2013multi} and construct different heads for each part of the date.
For the architecture that predicts both the day, month, and entire year, we construct seven classification heads in total:
Two heads to predict the date (one for the first and one for the second digit), another head to predict the month, and four heads to predict the year (one for each of the four digits).
As opposed to \citet{goodfellow2013multi}, we do not estimate the length of our sequences; while house numbers differ in lengths, the length of a date is not well-defined.\footnote{While part of a date can be missing, the length of a date is not well-defined, as the absence of, e.g., the month does not change the position of the remaining predictions: The day is still the day and the year is still the year. This is different for house numbers: If the estimated length of a house number is three, a non-missing prediction of a fourth number should be discarded.}
Further, the number of classes differs between our classification heads:
The first day classification head can predict four classes (1-3 and missing), the second day classification head can predict 11 classes (0-9 and missing), the month classification head can predict 14 classes (1-12, missing, and wildcard\footnote{We use this to predict the presence of a special ``same as above''-token in the dates from the Swedish Cause of Death Register.}), and all four year classification heads can predict 11 classes (0-9 and missing).\footnote{In most settings, the first two year classification heads can be made smaller by only allowing them to predict fewer classes. For example, it is rare that the first digit of a year is anything but 1, 2, or missing, and in many settings even the second digit is also constrained within fewer classes. However, if it is the case that these two first digits of the year are fixed, we recommend that the use of our model only predicting the day, month, and part of the year.}
For the network predicting the day, month, and last two digits of the year, we remove the other two heads related to the year (leading to a total of five classification heads).
For the network predicting the day and month, we remove all heads related to the year (leading to a total of three classification heads).

\subsection{Optimization}
We use the same basic recipe to train our networks, regardless of the specific architecture.
We use stochastic gradient descent with momentum as our optimizer and a cosine annealing learning rate scheduler with warmup to control the learning rate.
We regularize our networks using dropout \citep{srivastava2014dropout} and stochastic depth \citep{huang2016deep}.
We further employ adaptive gradient clipping \citep{brock2021high}, label smoothing \citep{szegedy2016rethinking}, and a small amount of weight decay.
We use image augmentation in the form of Random erase \citep{zhong2020random} and RandAugment \citep{cubuk2020randaugment}.

As our optimization criterion, we draw inspiration from \citet{goodfellow2013multi} and estimate losses separately for each token, after which we combine the token losses to obtain the total loss.
However, our approach incorporates label smoothing and allows for a different number of classes across classification heads.
Further, we normalize the loss by the sequence length, to obtain more comparable losses across models with different numbers of classification heads.
For a sequence of $t=1,\dots,T$ elements, each with $C_t$ distinct classes, we seek to minimize the loss represented by Equation \ref{eq:loss}. Here, $\alpha_t$ controls the degree of label smoothing, $T$ the number of classification heads, and $C_t$ the number of classes for classification head $t$, $t=1,\dots,T$.
For $\alpha_t$ = 0 for all $t$, this maximizes the probability of a sequence given the image, provided the different elements are (conditionally) independent.
If additionally $T=1$, this is the standard cross entropy loss commonly used in, e.g., image classification, with $C_1$ number of classes.

\begin{align*}
    \mathcal{L} &= \frac{1}{T} \sum_{t=1}^T \sum_{j=1}^{C_t} \log \hat{y}_{t, j} \left[(1 - \alpha_t)y_{t, j} + \frac{\alpha_t}{C_t}\right] \numberthis \label{eq:loss}
    \\
    y_{t} & \in \{0, 1\}^{C_t}: \text{Correct one-hot label for the \textit{t}'th element}
    \\
    \hat{y}_{t} & \in [0, 1]^{C_t}: \text{Normalized scores for the \textit{t}'th element}
    \\
    \alpha_t & \in [0, 1): \text{Smoothing associated with the \textit{t}'th element}
\end{align*}

Across all our models, we resize our images to a height of 160 pixels and a width of 352 pixels.\footnote{This keeps the aspect ratio of our resized images within reasonable values compared to the source images.}
As initial weights, we use the weights from EfficientNetV2-S pre-trained on ImageNet21k \citep{russakovsky2015imagenet}.
As a default, we use a batch size of 256 and a learning rate of 0.6, and train for a total of 250 epochs, where the first ten are used to warm up the learning rate.
However, for some of our models -- those using a particularly large number of observations - we change the batch size, learning rate, and number of epochs used to warm up the learning rate. More details on the specific parameters are presented in Table \ref{tab:params} in \ref{model:params}.

\subsection{Networks}
In total, we train 11 neural networks as part of the DARE system, each on a different subset of the DARE database; here, we use PyTorch Image Models \citep{rw2021timm}.
First, for each of the eight individual datasets (see Table~\ref{tab:database}), we train a network with classification heads according to the information available i.e., we train a network with three classification heads if only the day and month are available, five classification heads if the day, month, and the last two digits of the year are available, and seven classification heads if the the day, month, and entire (four digits) year are available.
The transcription performance of these models serve as baseline performance measures, and provide valuable information on (a) what performance the neural networks are able to achieve in the absence of the full DARE database and (b) what improvements the full or partial DARE database provides over using networks trained on individual datasets.
We denote those models \texttt{M-X}, where \texttt{X} is the dataset used for training (i.e., \texttt{M-DC-1} is the model trained on the first subset of the Danish death certificates).

Second, we train a neural network on all datasets of the DARE database, which predicts both the day, month, and (entire) year. When, for example, part of the year is not available, this network is trained to predict that part of the year as missing (i.e., it can still transcribe part of a year). We denote this model as \texttt{M-DDMYYYY}. 
This model's main strength is that it is trained on the largest possible sample, which also includes images from the highest number of different distributions, allowing the network to learn from the most diverse set of dates.
However, this introduces certain complications.
Often the year is present on an image without the label stating so (for the datasets with no or only partial year information), in which case the model is given quite an unfair task: Even if the year is present on an image, it is sometimes (without consistency) taught to predict it as missing, as that is the case according to the label.
Further, forcing one model to transcribe in very diverse settings, including settings where no information on the year is present on the images and settings where all four digits of the year are available, might be sub-optimal.

Our third set of models consists of two neural networks, both trained on subsets of the DARE database:
First, we train a network on the five datasets where at least some year information is available.
This network includes only five classification heads, i.e. it only transcribes two digits of the year.
In most cases, this is sufficient for any downstream analysis and it also avoids the issue discussed above, where discrepancies between the label and the image could degrade performance.
Second, we train a network on the remaining three datasets where only the day and month is available.
We denote the first of these models \texttt{M-DDMYY} and the second \texttt{M-DDM}.

\subsection{Performance measures}
To measure the performance of our networks, we focus on the sequence accuracy (SeqAcc) of our transcriptions.
A date prediction is deemed correct if all parts of the transcription is correct, and thus incorrect if any one of up to seven token predictions is incorrect; this also means that the token accuracies of our networks are significantly higher than the reported sequence accuracies.

\begin{table}[!ht]
    \centering
    \footnotesize{
            \caption{Transcription Performance}
            \label{tab: transcription-performance}
    \resizebox{1\textwidth}{!}{
        \begin{tabular}{lllrrrr}
            \toprule
            & & & & \multicolumn{3}{c}{\textbf{Decomposed Accuracy}} \\ \cline{5-7}
            \textbf{Dataset} & \textbf{Model} & \textbf{Sequence} & \textbf{SeqAcc} & \textbf{DayAcc} & \textbf{MonthAcc} & \textbf{YearAcc}\\
            \midrule
            \multirow{3}{*}{\textbf{DC-1}} & \texttt{M-DC-1} & DD-M-YY(YY) & {97.50}\% (97.30\%) & 99.20\% & 98.80\% & \textbf{99.40}\% (99.20\%) \\
            & \texttt{M-DDMYYYY} & DD-M-YY(YY) & \textbf{97.70}\% (96.70\%) & 99.20\% & \textbf{99.00}\% & \textbf{99.40}\% (98.20\%) \\
            & \texttt{M-DDMYY} & DD-M-YY & {97.60}\% & \textbf{99.30}\% & \textbf{99.00}\% & 99.20\% \\
            \hline
            \multirow{3}{*}{\textbf{DC-2}} & \texttt{M-DC-2} & DD-M-YY(YY) & \textbf{92.88}\% (92.78\%) & 97.75\% & \textbf{98.19}\% & \textbf{94.67}\% (94.58\%) \\
            & \texttt{M-DDMYYYY} & DD-M-YY(YY) & 91.71\% (91.59\%) & 97.78\% & 98.18\% & 93.49\% (93.36\%)\\
            & \texttt{M-DDMYY} & DD-M-YY & {91.92}\% & \textbf{97.88}\% & 98.18\% & 93.63\%\\
            \hline
            \multirow{3}{*}{\textbf{PR-1}} & \texttt{M-PR-1} & DD-M-YY & {93.47}\% & 97.21\% & 98.18\% & 97.45\% \\
            & \texttt{M-DDMYYYY} & DD-M-YY & 94.01\% & \textbf{97.43}\% & 98.31\% & \textbf{97.69}\% \\
            & \texttt{M-DDMYY} & DD-M-YY & \textbf{94.02}\% & 97.40\% & \textbf{98.32}\% & 97.68\% \\
            \hline
            \multirow{3}{*}{\textbf{PR-2}} & \texttt{M-PR-2} & DD-M-YY & 84.21\% & 89.12\% & 91.24\% &  90.86\% \\
            & \texttt{M-DDMYYYY} & DD-M-YY & \textbf{85.24}\% & 89.67\% & 91.76\% & \textbf{91.26}\% \\
            & \texttt{M-DDMYY} & DD-M-YY & {85.15}\% & \textbf{89.69}\% & \textbf{91.77}\% &  91.15\%\\
            \hline
            \multirow{3}{*}{\textbf{SWE-BD}} & \texttt{M-SWE-BD} & DD-M-YY & 94.49\% & {97.45}\% & 97.22\% & 96.59\% \\
            & \texttt{M-DDMYYYY} & DD-M-YY & \textbf{94.73}\%  & \textbf{97.58}\% & \textbf{97.36}\% & \textbf{96.75}\%\\
            & \texttt{M-DDMYY} & DD-M-YY & {94.68}\% & 97.52\% & 97.31\% & 96.73\% \\
            \hline
            \multirow{3}{*}{\textbf{FR}} & \texttt{M-FR} & DD-M & \textbf{98.24}\% & 99.21\% & \textbf{98.96}\% & \\
            & \texttt{M-DDMYYYY} & DD-M & 98.19\% & \textbf{99.27\%} & 98.84\% & \\
            & \texttt{M-DDM} & DD-M & \textbf{98.24}\% & 99.25\% & 98.91\% & \\
            \hline
            \multirow{3}{*}{\textbf{NHVD}} & \texttt{M-NHVD} & DD-M & \textbf{97.53}\% & \textbf{98.13}\% & 98.85\% & \\
            & \texttt{M-DDMYYYY} & DD-M & 97.48\% & 98.05\% & \textbf{98.95}\% & \\
            & \texttt{M-DDM} & DD-M & 97.52\% & 98.03\% & 98.91\% &\\ 
            \hline
            \multirow{3}{*}{\textbf{SWE-DD}} & \texttt{M-SWE-DD} & DD-M & {99.39}\% & 99.51\% & 99.70\% & \\
            & \texttt{M-DDMYYYY} & DD-M & 99.34\% & 99.49\% & 99.69\% &  \\
            & \texttt{M-DDM} & DD-M & \textbf{99.40}\%  & \textbf{99.53}\% & \textbf{99.72\%} &  \\
            \bottomrule
        \end{tabular}}
    }
    \begin{minipage}{1\linewidth}
        \vspace{1ex}
        \scriptsize{
            \textit{Notes:}
            \linespread{1.0}\selectfont
                The table shows the date transcription performance of our models measured on either the DD-M or DD-M-YY sequence.
                Note that the sequence used to evaluate the models differ in the case of the \textbf{DC-1} and \textbf{DC-2} datasets, as the \texttt{M-DDMYY} model only transcribes the last two digits of the year while the \texttt{M-DDMYYYY} and \texttt{M-DC-\{1, 2\}} models are able to predict the full date sequence.
                For a fair comparison we report both the partial as well as the full SeqAcc, with the full SeqAcc written in paranthesis.
                Table~\ref{tab:database} shows the number of observations of the test split for each dataset.
            }
    \end{minipage}
\end{table}

\section{Transcription performance}
\label{sec: transcript performance}
Table~\ref{tab: transcription-performance} shows the performance of our date transcription models for each individual dataset of the DARE database, with the best result for each dataset marked in bold.
Across all but one dataset ({PR-2}), all models reach high levels of SeqAccs, ranging from 92\% to more than 99\%. \footnote{The poor performance on {PR-2} stems from primarily two causes: First, the segmentation of these fields is quite poor, and thus the date is not always present on the segmented fields which is a common issue in this sequential segmentation and recognition pipeline \citep{Curtis2018}. Second, the label quality is poor, meaning that the label does not always reflect the date that is actually present on the image. For these reasons, the performance on this dataset does not reflect actual performance directly, as it is impossible to reach anywhere close to 100\% accuracy due to the poor quality of the data. Nonetheless, we find that the data is still usable for training, improving the generalization of our models.}
In comparison, \citet{dahl2021applications} found that Transkribus obtained a sequence accuracy of 74.9\% and 96.3\% using crowdsourcing on {DC-1}.
In comparison, our models achieve a SeqAcc of over 97\% on this subset of the Danish death certificates.
Hence, the DARE system does not only provide benefits in terms of cost and scale compared to manual labelling, but also increases the accuracy of the transcriptions.

\begin{figure}
    \centering
    \caption{Incorrect PR-1 Transcriptions \\
     (ground truth vs. transcription)}
    \label{fig:database-inccorect-PR}
    \subfloat[1-1-03 vs. 1-1-02]{
        \includegraphics[height=0.08\textwidth]{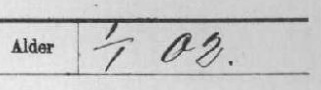}
        \label{fig:examples_pr_a}
    }
    \subfloat[1-9-92 vs. 1-9-82]{
        \includegraphics[height=0.08\textwidth]{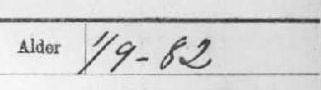}
        \label{fig:examples_pr_b}
    }
    \subfloat[2-7-91 vs. 21-7-88]{
        \includegraphics[height=0.08\textwidth]{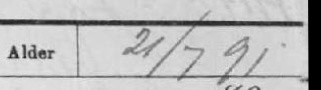}
        \label{fig:examples_pr_c}
    }
    \qquad
    \subfloat[2-9-73 vs. 3-9-73]{
        \includegraphics[height=0.08\textwidth]{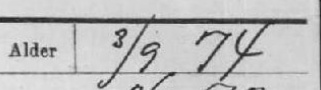}
        \label{fig:examples_pr_d}
    }
    \subfloat[5-10-93 vs. 3-10-93]{
        \includegraphics[height=0.08\textwidth]{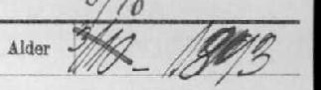}
        \label{fig:examples_pr_e}
    }
    \subfloat[3-1-47 vs. 5-1-47]{
        \includegraphics[height=0.08\textwidth]{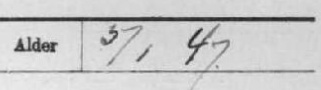}
        \label{fig:examples_pr_f}
    }
    \qquad
    \subfloat[3-1-89 vs. 3-1-87]{
        \includegraphics[height=0.08\textwidth]{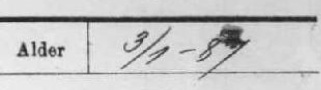}
        \label{fig:examples_pr_g}
    }
    \subfloat[3-4-07 vs. 3-11-07]{
        \includegraphics[height=0.08\textwidth]{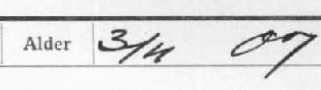}
        \label{fig:examples_pr_h}
    }
    \subfloat[4-6-68 vs. 4-6-66]{
        \includegraphics[height=0.08\textwidth]{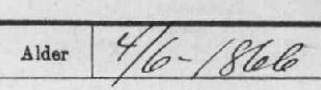}
        \label{fig:examples_pr_i}
    }
    \qquad
    
    \begin{minipage}{1\linewidth}
        \vspace{1ex}
        \scriptsize{
        \textit{Notes:}
        The figure shows examples of incorrectly transcribed police register sheets dates (ground truth vs. transcription).
        The images are a random subset of the test set where the M-DDMYY model incorrectly transcribed the dates, in the sense of disagreeing with the label.
        Note how some of these ground truth labels are also incorrect, specifically Panels \ref{fig:examples_pr_a}, \ref{fig:examples_pr_b}, \ref{fig:examples_pr_c}, \ref{fig:examples_pr_d}, \ref{fig:examples_pr_g}, and \ref{fig:examples_pr_i}.
        Further, it is difficult to judge whether the ground truth of the images in Panels \ref{fig:examples_pr_e} and \ref{fig:examples_pr_f} are correct.
        }
    \end{minipage}
\end{figure}

In Figure \ref{fig:database-inccorect-PR} we show some of the examples of incorrect transcriptions on the {PR-1} dataset according to the ground truth. As stated in \citet{goodfellow2013multi} the ground truth labels in real world datasets are often noisy, which is also the case in our setting here. Manual transcriptions are not perfect in the database which can be seen from e.g. Panel \ref{fig:examples_pr_a}, \ref{fig:examples_pr_b}, \ref{fig:examples_pr_c}, \ref{fig:examples_pr_d}, \ref{fig:examples_pr_g} and \ref{fig:examples_pr_i}, i.e., more than half of the randomly selected incorrect transcriptions have an incorrect ground truth and in three of these (Panel \ref{fig:examples_pr_a}, \ref{fig:examples_pr_g}, and \ref{fig:examples_pr_i}) the transcription seems to outperform the ground truth. It therefore seems likely that the reported SeqAccs are lower bounds for the actual transcription performance as both the manual and network transcriptions have to be in agreement and the SeqAcc does not take into account falsely labelled ground truth dates. Other documents are either cropped or written in an ambiguous manner, examples of these can be seen in Panel \ref{fig:examples_pr_e} and \ref{fig:examples_pr_f}. Panel \ref{fig:examples_pr_e} are cropped around the region of interest, but the date (day and month) seems to be crossed out and an updated date is written above the region of interest (looking at the original image and the ground truth the date written above the region of interest is 5-10). This is a cost from the segmentation of the documents as we are relying on the original writers to strictly stick to the document forms. Panel \ref{fig:examples_pr_f} is a different issue, since it is difficult to read the date due to the degradation of the day on the document (3 vs 5). Both 3 and 5 seem feasible and making a completely confident transcription is troublesome. It seems as though there is a very weak slightly darker area to the left of the top of the day, which could indicate that the date is indeed 3-1-47.

\section{Transfer learning}
\label{sec: tl}
While Section~\ref{sec: transcript performance} demonstrated the performance of the DARE system in an internal sense (achieving high transcription performance on the test splits of the datasets used to train the system), our ultimate goal for the DARE system is to show its generalizability to other datasets.
In this section, we document that using the DARE system for transfer learning leads to significant improvements in transcription accuracy of dates from historical Swedish grade sheets.
To this end, we train networks to transcribe dates from Swedish grade sheets with and without transfer learning from the DARE system, demonstrating that using the DARE system for transfer learning improves performance.
Further, we demonstrate that transfer learning from the DARE system is better than transfer learning from ImageNet21k \citep{russakovsky2015imagenet}, a dataset of 13 million images that is commonly used to pre-train models before finetuning on downstream tasks.
That transfer learning from the DARE database is significantly better is particularly promising, as pre-training on the ImageNet21k dataset is used to achieve the best performance in image classification, aside from the use of non-public Instagram/JFT images. We train eight neural networks to transcribe these grade sheets:
One trained from scratch, one using ImageNet21k pre-trained weights (similar to how we train the networks in Section~\ref{sec: DARE System}), and one using DARE pre-trained weights.
We use largely the same architecture and training recipe as discussed in Section~\ref{sec: DARE System} to train these networks.
As the century is rarely available on the image and never differs when available (only ever being 19) we only transfer from models trained on a sequence containing some year information, using the architecture with five heads (\texttt{M-DDMYY}).
When training our networks, we use the same recipe with the exception of the learning rate and number of epochs.\footnote{Further, we resize images to 224 by 224, for a roughly equal number of pixels without changing the aspect ratio of the source images.}
Due to the difference in training set size and the use of transfer learning for seven of the eight models, we perform a small search over exponentially distanced learning rates for each network.
Further, we experiment with a longer training schedule in terms of number of epochs (due to the small training set size).
We then use the learning rate and number of epochs that perform best on the 2,000 training images and finally evaluate the performance of the models on the test set of 4,139 images.

\begin{table}[!ht]
    \centering
            \caption{Transcription Performance: Swedish Grade Sheets}
            \label{tab: tl-atlass}
%    \resizebox{1\textwidth}{!}{
    \footnotesize{
        \begin{tabular}{llrrr}
            \toprule
                \textbf{Pretrained} & \textbf{Sequence} &  \textbf{SeqAcc} &  \textbf{TL Gain} &  \textbf{Error Rate Reduction} \\
                \midrule
                No & DD-M-YY & 85.17\% &  \\
                ImageNet21k & DD-M-YY & 93.38\% & +8.21\% & -55.37\% \\
                \texttt{M-DC1} & DD-M-YY & {95.19}\% & +{10.03}\% & -{67.59}\% \\
                \texttt{M-PR2} & DD-M-YY & {95.39}\% & +{10.22}\% & -{68.89}\% \\
                \texttt{M-DC2} & DD-M-YY & {95.63}\% & +{10.46}\% & -{70.52}\% \\
                \texttt{M-SWE-BD} & DD-M-YY & {95.63}\% & +{10.46}\% & -{70.52}\% \\
                \texttt{M-PR1} & DD-M-YY & {95.89}\% & +{10.73}\% & -{72.31}\% \\
                \texttt{M-DDMYY} & DD-M-YY & \textbf{96.18}\% & +\textbf{11.02}\% & -\textbf{74.27}\% \\
                \bottomrule
            \end{tabular}}
%            }
            \begin{minipage}{1\linewidth}
            \linespread{1.0}\selectfont
                \vspace{1ex}
                \scriptsize{
                \textit{Notes:}
                The table shows the SeqAcc on the Swedish grade sheets of eight different models:
                One trained from scratch, one finetuned from ImageNet21k pre-trained weights, and six finetuned from DARE pre-trained weights.
                It also shows the SeqAcc improvement and the error rate reduction obtained by using transfer learning, compared to the baseline model trained from scratch. Rows are ordered according the their imporovement over the baseline model.
                }
            \end{minipage}
\end{table}

Table~\ref{tab: tl-atlass} shows the SeqAcc of all eight models as well as the SeqAcc improvement and the error rate reduction obtained by using transfer learning compared to the model trained from scratch. 
There is significant variation in the transcription performance achieved by the different models, with the SeqAcc ranging from 85.17\% for the model trained from scratch to 96.18\% for the best model transfer learning from the DARE system. We find that transfer learning from the full database is superior to transfer learning from subsets of the DARE database.  While {PR-2} is the second largest dataset, transfer learning from this dataset is outperformed by two smaller dataset ({DC-2} and {SWE-BD}). We conclude from this that it is not only the size of the transfer learning dataset that is important, but rather the interaction between quality and quantity. 
Hence, the best performance is achieved when transfer learning from the \texttt{M-DDMYY} model -- outperforming not only training from scratch but also transfer learning from smaller subsets of the DARE database or ImageNet21k.
This demonstrates the usefulness of the DARE system:
At no cost in required training resources or time, date transcription accuracy is significantly improved on this new dataset. In fact, in order to obtain the results for the model trained from scratch, a much longer training schedule was used, meaning that transfer learning reduced the computational resources required to train the model.

\begin{figure}
    % \captionsetup{width=.9\linewidth}
    \caption{Performance on Swedish Grade Sheets}
    \label{fig: tl-atlass}
    \centering
    \includegraphics[width=0.9\linewidth]{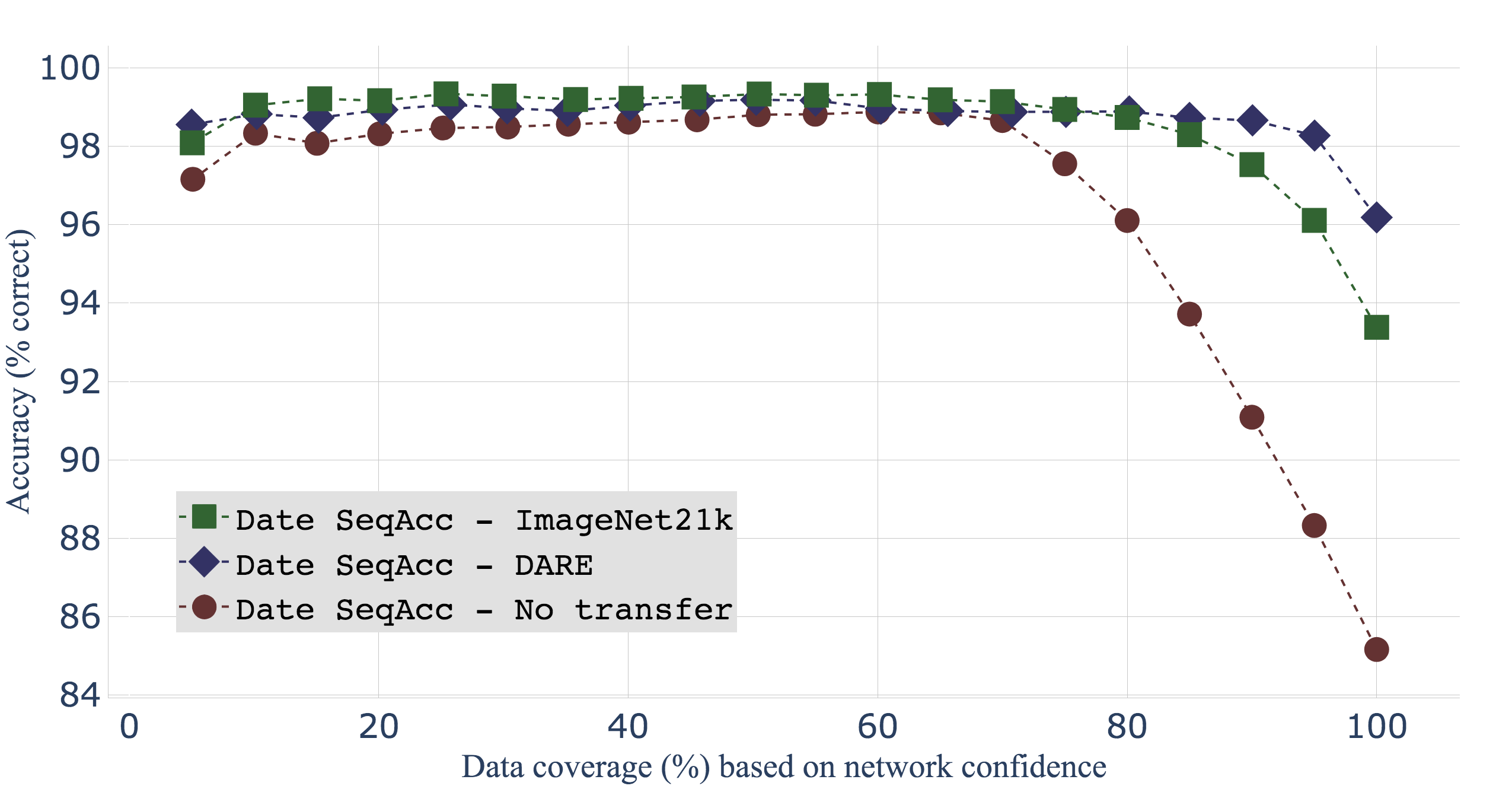}
    \begin{minipage}{1\linewidth}
    \linespread{1.0}\selectfont
        \vspace{1ex}
        \scriptsize{
        \textit{Notes:}
        The figure shows the accuracy at different levels of data coverage on the Swedish grade sheet data for three models:
        One trained from scratch, one using ImageNet21k pre-trained weights, and one using DARE (here represented by M-DDMYY) pre-trained weights.
        The models are evaluated on between 5\% and 100\% data coverage, meaning that the leftmost points are measures of the accuracy if we keep only the 5\% of predictions that the network is most confident about and the rightmost points are the SeqAccs. 
%        \co{TODO: Change DARE to \texttt{M-DDMMYY} now where we show for multiple DARE models. Only remember to change more generally text where we refer to the DARE model to now correctly state that it is the \texttt{M-DDMMYY} model we TL from on those cases.}
        }
    \end{minipage}
\end{figure}

\begin{figure}
    % \captionsetup{width=.9\linewidth}
    \caption{Error Rate Reduction on Swedish Grade Sheets}
    \label{fig: error_rate}
    \centering
    \includegraphics[width=0.9\linewidth]{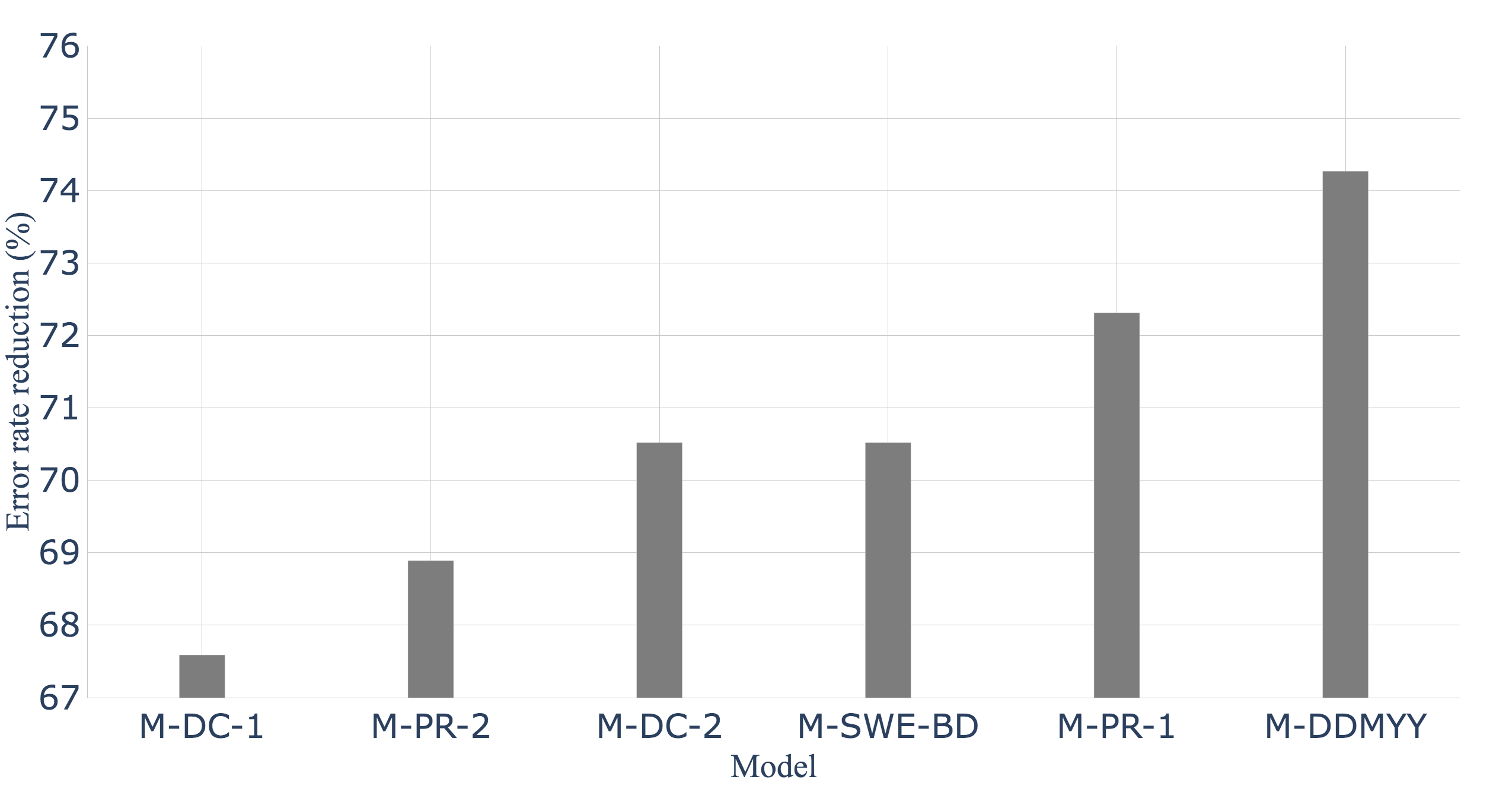}
    \begin{minipage}{1\linewidth}
    \linespread{1.0}\selectfont
        \vspace{1ex}
        \scriptsize{
        \textit{Notes:}
        The figure shows the error rate reduction achieved by transfer learning from six different models from the DARE system (the M-DDMYY model), compared to training from scratch, on Swedish grade sheets.
        The error rate reduction is calculated as the relative reduction in SeqAcc achieved by the models compared to the baseline model (no transfer learning).
        The models are ordered according to the error rate reduction they achieve, such that the rightmost bar corresponds to the model that achieves the largest error rate reduction.
        }
    \end{minipage}
\end{figure}

Another useful feature of our date transcription model is presented in Figure~\ref{fig: tl-atlass}, which shows the SeqAcc of three models (no transfer learning, ImageNet21k, and \texttt{M-DDMYY}) at different levels of data coverage.
Specifically, we gradually remove predictions where the models are the least certain and evaluate the models on the remaining images.
Similarly for all models, removing 10 to 20\% of images where they are the least certain significantly improves their performance on the remaining set.
This shows that all the models are able to identify the subset of images where they might need assistance, and the large subset where our confidence in their performance is high.
Further, it shows that achieving, e.g., 98\% SeqAcc can be achieved at a much higher data coverage for the model transfer learning from the DARE database compared to the other models.
More broadly, the figure shows that the models all perform fairly equally up until 70\% data coverage. 
From this point on the model trained without transfer learning experiences a sizable fall in its performance.
We find a similar pattern at around 80\% data coverage for the model using the pre-trained weights from ImageNet21k.
To achieve, e.g., 98\% SeqAcc at above 80\% data coverage, only the model transfer learning from the DARE database is adequate and able to retain this high accuracy until around 95\% data coverage.
Further, the approach we use is completely generalizable to all settings where segmented fields of dates require transcription -- no part of this is made specifically for the Swedish grade sheets.
This allows others to directly use the DARE system for their own projects on any dataset and potentially achieve significantly improved transcription performance. To further showcase the importance of the combined DARE database we illustrate the respective performance achieved by transfer learning from only parts of the database relative to the full \texttt{M-DDMYY} model in Figure \ref{fig: error_rate}. 

Figure \ref{fig: error_rate} shows the error rate reduction achieved by transfer learning from six different models. We find that there seems to be a connection between the size (and quality) of the dataset used for transfer learning since these are sorted according to the accuracy from left to right and this roughly corresponds to the size of the different datasets. The exception to this is \texttt{M-PR-2} which performs worse than \texttt{M-DC-2} and \texttt{M-SWE-BD} (note that {SWE-BD} ``only'' consists of 198,096 transcribed dates). We believe this likely is caused by the relatively poor quality of the {PR-2} dataset. In addition, we find that transfer learning from the combined DARE database outperform the performance relative to the other models trained on only a single dataset. Although, the absolute performance in terms of accuracy ``only'' ranges from 95.19\% to 96.18\% (see Table \ref{tab: tl-atlass}) we find that there is an advantage from transfer learning from several and larger datasets.

% -----------------------------------------

\section{Linking pipeline}
\label{sec: linking}
To illustrate another important application of the DARE system, we use it for linking and matching more than 140,000 individuals. Originally, information on these individuals has been manually transcribed by crowd sourcing but links back to the source images was not part of the transcription. Below we discuss how to link the manual transcriptions to the source images and subsequently we describe how to use the linked entries for training and transcribing a total of 3.7 million entries from the 1916 Danish Census, without requiring additional manual transcriptions. 

%\subsection{Linking pipeline}
As a starting point, we obtained 156,712 manually transcribed census records from the National Archives covering parts of the population of Denmark in 1916.
As already mentioned, the manual transcriptions are not linked to the scanned records. They only contain a reference to the folder from which they were extracted. 
Therefore, in order to use the transcriptions for training, we have to link the scanned documents to the transcriptions, using matching. 
We suggest to use a sequential matching procedure which in many ways are similar to a linking pipeline when combining corpuses across data sources. 
In addition, we avoid manual labelling by implementing a pretrained date network, namely the \texttt{M-DDMYY} model. %\footnote{\co{Although almost one hour was spent on manually labelling empty cells from the Danish census. This was to avoid empty cells being falsely matched to the transcribed records.}}, 
While the date alone is not enough to identify individuals (due to many individuals with the same birth date), using the first and last name in combination with the date allows us to identify most individuals uniquely (in combination with information on the reference to the source folder). We therefore also transcribe the names (first and last) of the individuals in the census, using networks from \citet{dahl2022hana}.
%Our proposed linking pipeline works by first ``bootstrapping'' the process by directly using transcriptions of dates and names from, respectively, the \texttt{M-DDMYY} model of the DARE system and a first and a last name transcription model from \citet{dahl2022hana}.

\begin{figure}
    % \captionsetup{width=1\linewidth}
    \caption{Pipeline for Linking to the Danish Census}
    \label{fig:linking}
    \centering
    \begin{tikzpicture}[scale=4]
        \draw[dotted](-0.5,0.5) -- (2.5,0.5);
        \draw[dotted](2.5,0.5) -- (2.5,2.5) ;
        \draw[dotted](2.5,0.5) -- (2.5,2.5);
        %  \draw[dotted](2.5,0.5) -- (2.5,2.5) node [midway, right] {\Large\textbf{{x5}}} ;
        %  \draw node  to [out=330,in=300,looseness=8] (3);
        \draw[dotted](2.5,2.5) -- (-0.5,2.5);
        \draw[dotted](-0.5,2.5) -- (-0.5,0.5);
        %    \GraphInit[vstyle=Classic]
        \SetUpVertex[Lpos=-90]
        %      \tikzset{VertexStyle/.style = {shape=ellipse, fill=black,
        %                           minimum size=13pt,inner sep=0pt}
        %      }
        \SetGraphUnit{3} 
        \node[circle,text=black, fill =lightgray,text opacity = 1,fill opacity=0.00] (x 5) at (2.6,1.5) {\textbf{x5}};
        %\tikzset{VertexStyle/.style = {shape = oval,minimum size = 4pt}}
        \node[ellipse,text=black, fill =lightgray,text opacity = 1,fill opacity=0.1] (a) at (1,3) {Start};
        \node[ellipse,text=black, fill =brown,text opacity = 1,fill opacity=0.1] (b) at (1,2) {Predict};
        \node[ellipse,text=black, fill =cyan,text opacity = 1,fill opacity=0.05] (c) at (2,1.5) {Match};
        \node[ellipse,text=black, fill =teal,text opacity = 1,fill opacity=0.05] (d) at (1,1) {Extract};
        \node[ellipse,text=black, fill =green,text opacity = 1,fill opacity=0.05] (e) at (0,1.5) {Transfer};
        \tikzset{EdgeStyle/.style={->,font=\scriptsize},{below=15pt}}
        \Edge[label = Step 0](a)(b) 
        \Edge[label = Step 1](b)(c) 
        \Edges[label = Step 2](c,d)
        \Edges[label = Step 3](d,e)
        \Edges[label = Step 4](e,b)
        %         \GraphInit[vstyle=Classic]
    \end{tikzpicture}
    \begin{minipage}{1\linewidth}
    \linespread{1.0}\selectfont
        \vspace{1ex}
        \scriptsize{
        \textit{Notes:}
        The figure shows the process for linking and matching to the Danish census data. 
        Initially we predict on the full census data from 1916. 
        We match the predictions to the manually transcribed records. 
        We match exactly by region and two (or more) other identifiers (first name, last name, and date). After the initial round we match our predictions twice each round, one time using the predictions obtained from the first set of models and once using the predictions from the second set of models. Subsequently, we only use the intersection of the two matched samples obtained from the two sets of models.
        Once the predictions are matched to the transcribed records, we extract these matches and create two new tuning datasets (each containing 50\% of the intersected matched entries) that can be used for transfer learning. 
        From the transfer learned models we perform the final step of predicting on all census records from 1916. 
        We define these steps as a single round, which we in turn do a total of five times before obtaining the final transcriptions.
        }
    \end{minipage}
\end{figure}
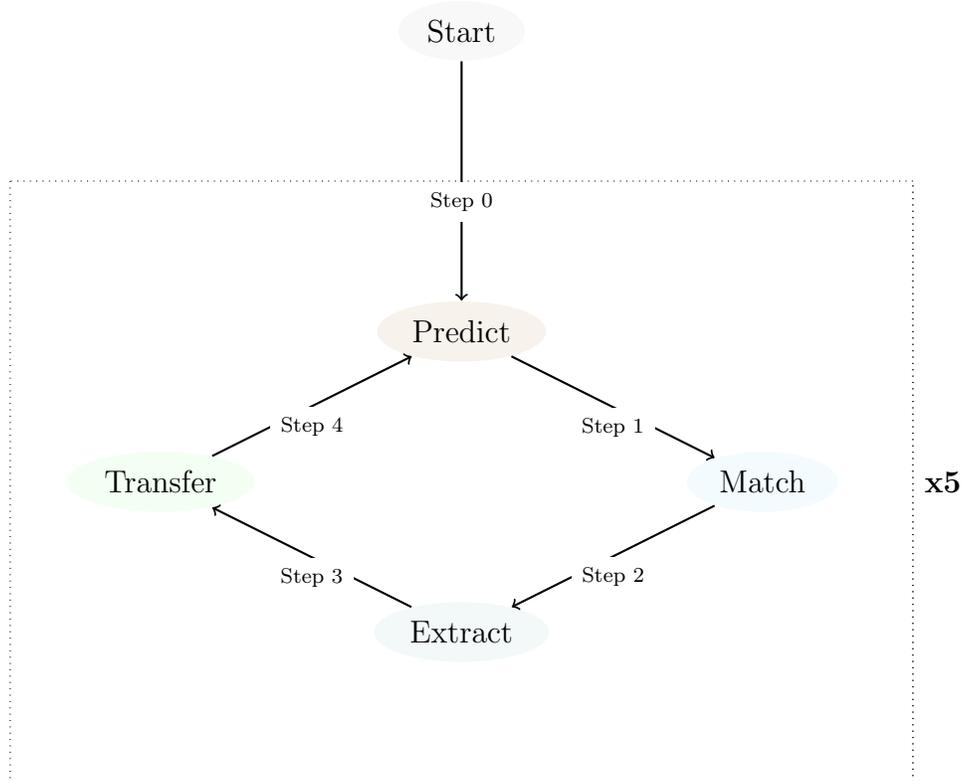

Our proposed linking pipeline use the DARE system to obtain date predictions for all dates from the 1916 Danish Census. We then match predictions that fulfill the criterion of having at least two equal (prediction, transcription)-pairs within the specific regions.
Specifically, we match an image to a transcription if at least two of three of the date, first name, and last name match exactly and there are no duplicate cases. %After ``bootstrapping'' (which we denote as the 0th round) our procedure, we perform several rounds (1-5) of iteratively matching additional images to the transcribed data., to gradually build up a larger training sample.
After matching, we use the matched data as a training set, and transfer learn from the three models (date, first name, last name). These models are used for predicting all entries from the 1916 census and we can repeat the matching process with the (expected) improved predictions. We perform several (5) rounds of iteratively predicting, matching and training to obtain additional linked images to the transcribed data.
Each of these rounds is similar, differing only with respect to what training data is available for use. However, we are faced with one important issue:
If we transcribe an image \textit{incorrectly}, but this incorrect transcription leads to a match, our networks may, due to simply memorizing the training data, keep transcribing this image incorrectly. To combat this issue, we need a process in our matching procedure to ``push'' out such cases, even if they enter our training data at some point.
To do so, we split the training sample into two and train networks separately on each of the two sets (that is, we train a total of six networks, two for each set, in each round).
We then use each set of networks to transcribe and match all images, and only keep the \textit{intersect} of the matched entries between the two sets of networks.
If an incorrect transcription enter our training data, this image will only be used to train one set of networks, and it is unlikely that the other set should transcribe (and match) this image incorrectly in the exact same way as the other set, and the incorrect transcription will thereby be pushed out of our training data. Figure~\ref{fig:linking} graphically illustrates this pipeline \footnote{Note that Figure~\ref{fig:linking} does not explicitly show how we train two sets of networks in each round and only train the following models on the intersection of the matched entries.}.

% We match predictions that fulfill the criteria of having at least two equal (prediction, transcription)-pairs within the specific regions which were extracted from meta information. 
% This equals almost 8,000 links in the initial round. We split these observations in order to perform cross-validation and majority voting for defining the labels for the following models. 
% Hence, based on 50\% of the 8,000 matches we train model 1.1 and the other half we use to train model 2.1. 
% Also, to transcribe the names we transfer learn the name network from \citet{dahl2022hana}. 
% The tuned networks are used for predicting on the full data once again, where we use the intersection of model 1.1 and 2.1 to define the labels for the following round. 
% We use only the intersection of the matched data and not the intersection of the predictions themselves. 
% This procedure is partially visualized in Figure \ref{fig:linking} but more extensively in the appendix in Figure \ref{fig:linking_expanded}.

Only in round 0 do we perform the 0th step, and for rounds 1-5 we go through the steps 1-4. After round 0, the steps are performed as:
\begin{enumerate}
    \item[] \textbf{Step 1}: Using predictions, match transcribed images to the transcribed data. Using only the intersection of the matched sample from the two sets of models.
    \item[] \textbf{Step 2}: Extract data from the intersection of the matched and transcribed images to use as a training sample. Split sample into two sets with respectively 50\% in each set.
    \item[] \textbf{Step 3}: For both subsets, train a date, a first name, and a last name transcription network, transfer learning from the \texttt{M-DDMYY} model for the date transcription and the models from \citet{dahl2022hana} for the name transcription.
    \item[] \textbf{Step 4}: Use the trained models to predict the dates and names of all images, resulting in two transcriptions for each image.
\end{enumerate}

In the setting of the 1916 Danish Census, the initial round leads to almost 8,000 matches.
We split these observations into two sets and train models based on the matched transcriptions.
Hence, based on 50\% of the 8,000 matches we train three models, one for date, one for first name, and one for last name) and the other half we use to train three other models.
These two sets of models are after training used to predict all images once again, and we use the intersection of the matches obtained from these two sets of models, to define the labels for the following round. 
As we repeat this process for several rounds we obtain more than 143,000 matched observations (out of the 156,712 potential matches).
We use this to train the final models that uses 100\% of the available training data and predict the entire Danish census.
These predictions constitute our final transcriptions, where we manage to match 146,081 of the 156,712 observations (a match rate of over 93\%).

\begin{figure}
    \centering
    \caption{Human Unreadable Dates from the 1916 Danish Census}
    \label{fig:census-unreadable}
    \subfloat[Date: 19-4-82 ; Name: Jens Koostrup]{
        \includegraphics[width=0.48\textwidth]{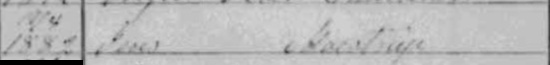}
        \label{fig:examples_census_a}
    }
    \subfloat[Date: 27-9-49 ; Name: Else Pommer]{
        \includegraphics[width=0.48\textwidth]{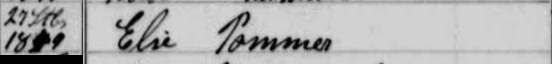}
        \label{fig:examples_census_b}
    }
    \qquad

    \subfloat[Date: 12-9-98 ; Name: Axel Schøle]{
        \includegraphics[width=0.48\textwidth]{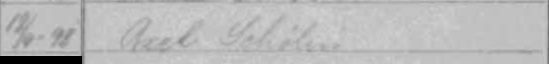}
        \label{fig:examples_census_c}
    }
\subfloat[Date: 18-1-02 ; Name: Karen Hansen]{
        \includegraphics[width=0.48\textwidth]{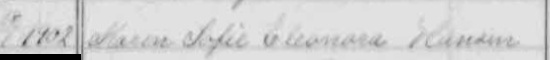}
        \label{fig:examples_census_d}
    }
    \qquad
    \subfloat[Date: 14-6-72 ; Name: Niels Pedersen]{
        \includegraphics[width=0.48\textwidth]{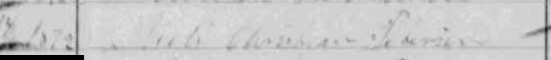}
        \label{fig:examples_census_e}
    }
    \subfloat[Date: 18-12-93 ; Name: Magda Mikkelsen]{
        \includegraphics[width=0.48\textwidth]{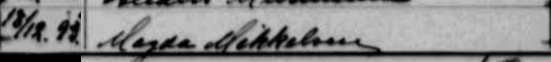}
        \label{fig:examples_census_f}
    }
    \qquad

    \begin{minipage}{1\linewidth}
    \linespread{1.0}\selectfont
        \vspace{1ex}
        \scriptsize{
        \textit{Notes:}
        This figure illustrates some of the examples where the DARE system transcriptions of the date were difficult to validate and we decided that we were not confident in labelling the transcription of the date as either correct or incorrect. For Panel \ref{fig:examples_census_a} and \ref{fig:examples_census_e} we were unsure about all transcriptions. For Panel \ref{fig:examples_census_d} and \ref{fig:examples_census_f} it is only the date transcription and for Panel  \ref{fig:examples_census_b} and \ref{fig:examples_census_c} we are unsure about the last name transcription as well.  
        }
    \end{minipage}
\end{figure}

%Appendix Figure~\ref{fig:linking_expanded} shows in detail how many observations each set of models are able to match for each round, including the number of observations in the intersection between the two sets of models for each round.  
While our matching procedure attempts to remove any cases of incorrect transcriptions, we perform a test to verify that the final transcriptions are correct.
Specifically, we manually review transcriptions of 1,000 randomly selected observations that were \textit{not} used as training data in the final round.
For both the date, first name, and last name, we manually compare the transcription to the source image, noting down whether (1) it is correct and (2) whether we are able to read the source image. Statement (2) is also illustrated in Figure \ref{fig:census-unreadable} where we provide some transcriptions where especially the date were difficult to validate and we decided that we were not confident in labelling the transcription of the date as either correct or incorrect.

Due to this complication, we define the projected SeqAcc as the SeqAcc on the sample we are able to read, as well as upper and lower bounds (defined, respectively, by labelling all illegible cases as correct or incorrect).
Table~\ref{tab: transcription-census} shows these SeqAccs.

\begin{table}[!ht]
    \centering
        \caption{Transcription Performance: 1916 Danish Census}
        \label{tab: transcription-census}
    \footnotesize{
%    \resizebox{1\textwidth}{!}{
        \begin{tabular}{lrrr}
            \toprule
        \textbf{Region} & \textbf{Lower Bound} & \textbf{Upper Bound} & \textbf{Projected}  \\
        \midrule
        \textbf{Date} & 90.70\% & 95.70\% & 95.47\% \\
        \textbf{First name} & 92.60\% & 95.30\% & 95.17\%  \\
        \textbf{Last name}  & 83.00\% & 88.10\% & 87.46\% \\
        \bottomrule
        \end{tabular}}
        \begin{minipage}{1\linewidth}
        \linespread{1.0}\selectfont
            \vspace{1ex}
            \scriptsize{
            \textit{Notes:}
            The table shows the manually evaluated transcription performance on the 1916 Danish Census. 
            We report three measures for each observation, as some images are unreadable and we cannot with certainty label these. 
            Hence, the lower bound is if we assume that all unreadable images are incorrectly predicted, the upper bound is if the unreadable images are all correctly predicted, and the projected is if the network obtains the same accuracy on the unreadable images as it does on the readable images which likely is an overestimation of the true expected performance.
            }
        \end{minipage}
%    }
\end{table}

Both the date and the first name are transcribed with high accuracy, with a a projected SeqAcc of respectively 95.47\% and 95.17\%. The projected SeqAcc can also be interpreted as the accuracy obtained if the network obtained the same transcription accuracy on the unreadable images as it does on the readable images. Presumably the true SeqAcc is somewhere between the projected and the lower bound, since the unreadable images supposedly are more difficult to transcribe for the network. The performance of the last name transcriptions is in general poorer, reflecting, in part, cases where no last name were stated. This unfortunately occurs quite frequently since the census records are sorted by families and a notation for ``same as above'' are quite frequently used for the last name. Even so, the projected accuracy is 87.46\% for the last name.

%% file: tex/4_discussion_and_conclusion.tex
\section{Discussion and Conclusion}
\label{sec: discussion-and-conclusion}
This paper introduces the DARE database and the DARE system.
The DARE database is the largest publicly available database of handwritten dates and consists of almost 10 million tokens from  3.1 million images with more than 2.2 million handwritten dates. We advocate that larger and more challenging publicly available datasets, together with the development of better and more efficient architectures for HTR models, will aid in overcoming the challenge of HTR for historical documents.  
%The database is based on different data sources from Denmark and Sweden extracted from tabulated documents.
Using this novel database, we train a number of neural networks able to transcribe dates -- the DARE system.
Similar to other authors \citep{Puigcerver2017, Kang2022} we do not use recurrent architectures and adopt a computationally more efficient model which is still able to obtain high accuracy.
The DARE system consists of a number of networks that differ with respect to which subsample of the DARE database they are trained on and what parts of a date they are trained to transcribe.
These networks achieve transcription accuracies in the range of 92\% to above 99\% on all but one dataset of the DARE database (the one exception being due to poor data quality).
Comparing the transcription accuracy on one of the datasets to that achieved through crowdsourcing and Transkribus, we find that we achieve more accurate date transcriptions.

While it is reassuring that the DARE system achieves high performance on the test splits of the datasets we use for training, our ultimate goal is to build a system that generalizes to completely new settings, which would prove useful for any researcher transcribing large historical archives. 
To demonstrate that this is the case for the DARE system, we first show that using the DARE system for transfer learning significantly improves transcription accuracy of dates from Swedish grade sheets (see Section~\ref{sec: tl}).
Further, even when comparing with a model finetuned from ImageNet21k pre-trained weights (or from smaller subsets of the DARE database), the model defined as M-DDMYY, finetuned from a larger part of the DARE database, still result in significantly more accurate transcriptions.

To illustrate another application for the DARE system we undertook the task of transcribing the name and date fields of the 1916 Danish Census with more than 3.7M observations in Section~\ref{sec: linking}.
Even though some of this data were already labeled, they were missing links to their image counterparts and thus no (image, label)-pairs were available for training.
To facilitate the transcription of the census, we first use the DARE system in a zero-shot way to transcribe dates in combination with name transcription models from \citet{dahl2022hana} in order to bootstrap a linking process between the labelled data and the images.
Directly using the predictions of our networks, we were able to link a small subset of the labels and images.
We used this to start an iterative process of gradually training better models and linking more images and labels; using transfer learning from the DARE system.
Ultimately we were able to train models that can transcribe the name and date fields of the 1916 Danish Census (excluding Copenhagen) in its entirety.\footnote{Further, with the link between images and the subset of labelled data made, it is simple to use the labelled data to train models to transcribe the remaining fields.}
This illustrates that the DARE system is not only useful for transfer learning:
It also has direct zero-shot use and it can be used to obtain vital information to link individuals across, for example, censuses.

%Both the transfer learning and linking procedure can be adopted into any other setting requiring transcription of dates. The system or database can therefore be directly used for increasing transcription accuracy in other settings extracting and transcribing dates from historical manuscripts. In addition, 
For future work, there are several dimensions along which the DARE system could be extended and improved. First, the DARE database could potentially benefit from including more datasets of handwritten dates. Also the quality of the datasets, i.e., the image snippets, could potentially be improved. For example, the segmentation process (producing image snippets) rely on the tabulated forms to be consistent and while this method is largely successful, it is not without error, and improving the segmentation of the fields would in all likelihood lead to a better database.
% consider to site Sanchez2019 here

Finally, the models of the DARE system should be maintained and updated over time. Whenever the DARE database is extended, re-training models on the larger set of datasets will be beneficial.
In addition, over time new architectures and new optimization strategies will be emerging. They must be evaluated and potentially (eventually) replace the existing models and strategies.

%Finally, other date formats might be of interest, for example, maybe only the year needs to be transcribed. Fortunately, it is straightforward and easy to augment the current DARE system with additional date models, i.e, \texttt{YYYY} and/or \texttt{YY} models.

%% file: tex/7_Appendix.tex
 \section{Details on Data Sources}
\label{sec: Details on Data Sources}
% In this appendix, we provide additional visualizations on the datasets we use to construct the DARE database.

%\subsection{Summary of the Final Database}

%The figure shows three examples of the date August 28, 33, each from a different dataset.
%Here, Panel~\ref{fig:examplesDC-1-33} shows an example where the day is written leftmost, the month in the middle but as text rather than a number, and the year rightmost; Panel~\ref{fig:examplesPR-1-33} differs in the sense that while the order of the components remains the same, the month is now written as a number; and Panel~\ref{fig:examplesSWE-BD-33} differs in the sense that now the year is written leftmost with the month still being written as a number.

%\section{The DARE Database}
%\label{sec: Constructing the DARE Database}
In this section, we first describe the construction of the DARE database, including the data sources used and how we extract the dates from the source images. 
% We then discuss the format of the dates and how they vary across the different data sources. %, as the information present on each differs. 
We then introduce two new datasets, the first dataset (Swedish grade sheets) is used directly in our transfer learning example and the second (1916 Danish Census) we use in combination with our linking method to match the images to a small labelled subset of the census. Finally, we provide additional visualizations of the datasets used to construct the DARE database.

\subsection{Datasets of the DARE Database}
\begin{figure*}
    \centering
    \caption{Date Images from the DARE Database}
    \label{fig: database-image-examples}
    \subfloat[Death Certificates (DC-1)]{
        \includegraphics[width=0.9\textwidth]{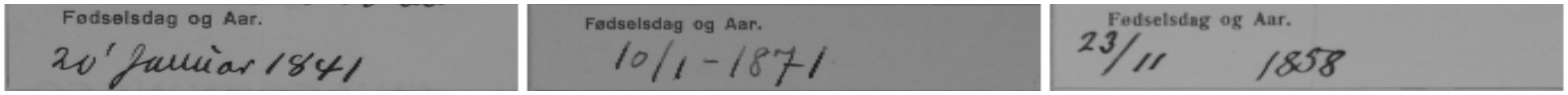}
        \label{fig:examplesDC-1}
    }
    \qquad
    \subfloat[Death Certificates (DC-2)]{
        \includegraphics[width=0.9\textwidth]{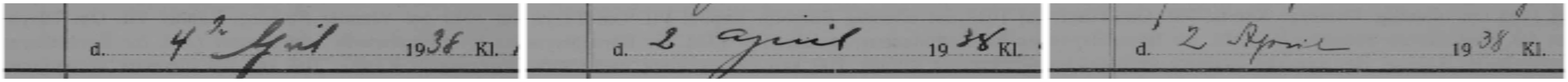}
        \label{fig:examplesDC-2}
    }
    \qquad
    \subfloat[Police Register - Main (PR-1)]{
        \includegraphics[width=0.9\textwidth]{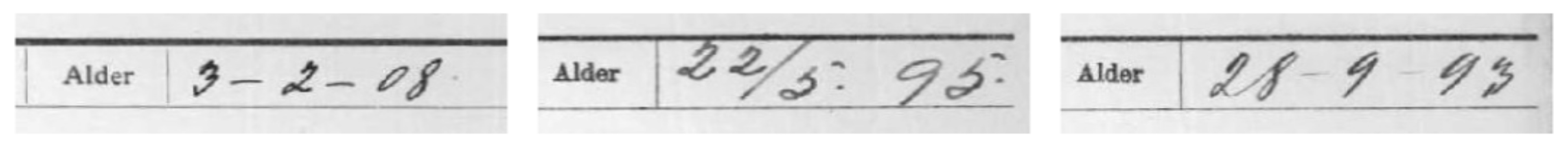}
        \label{fig:examplesPR-1}
    }
    \qquad
    \subfloat[Police Register - Spouse (PR-2)]{
        \includegraphics[width=0.9\textwidth]{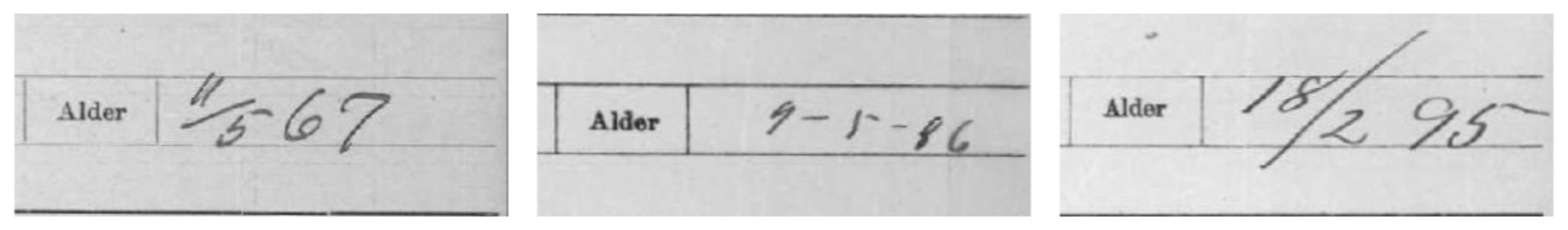}
        \label{fig:examplesPR-2}
    }
    \qquad
    \subfloat[Police Register - Child (PR-2)]{
        \includegraphics[width=0.9\textwidth]{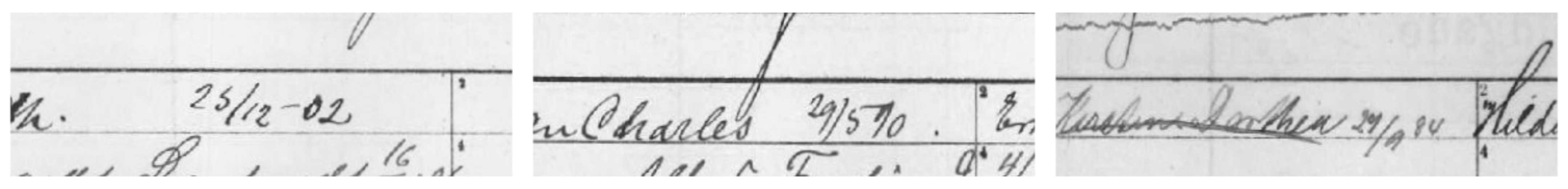}
        \label{fig:examplesPR-2-1}
    }
    \qquad
    \subfloat[Funeral Records (FR)]{
        \includegraphics[width=0.4\textwidth]{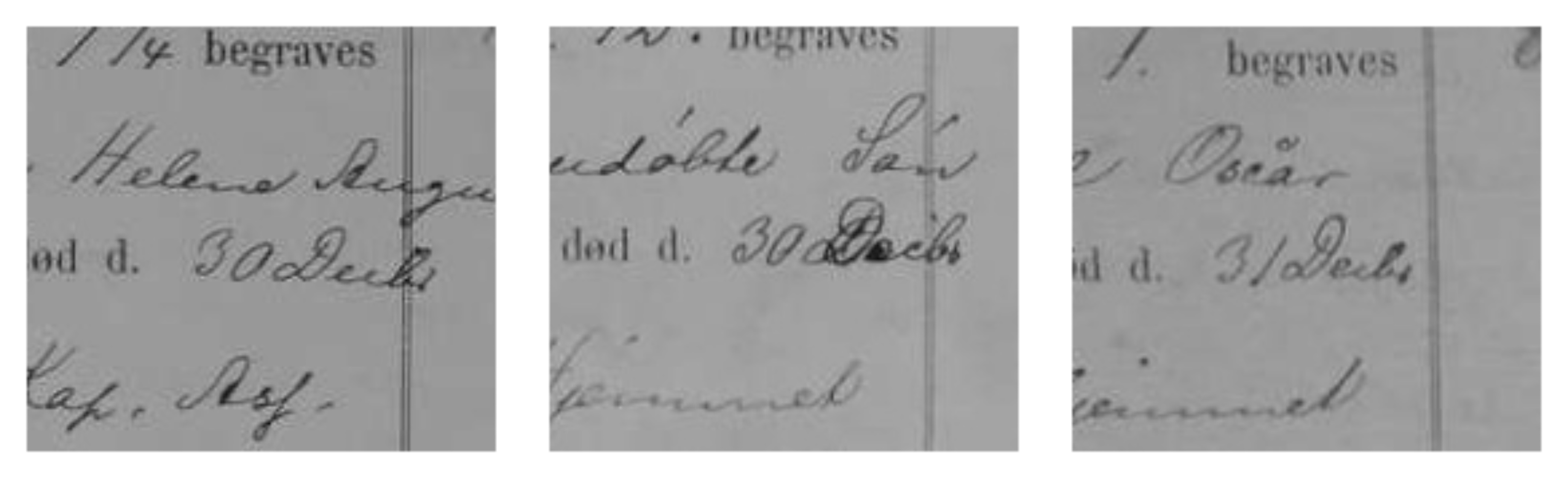}
        \label{fig:examplesFR}
    }
    \qquad
    \subfloat[Swedish Birth Dates (SWE-BD)]{
        \includegraphics[width=0.4\textwidth]{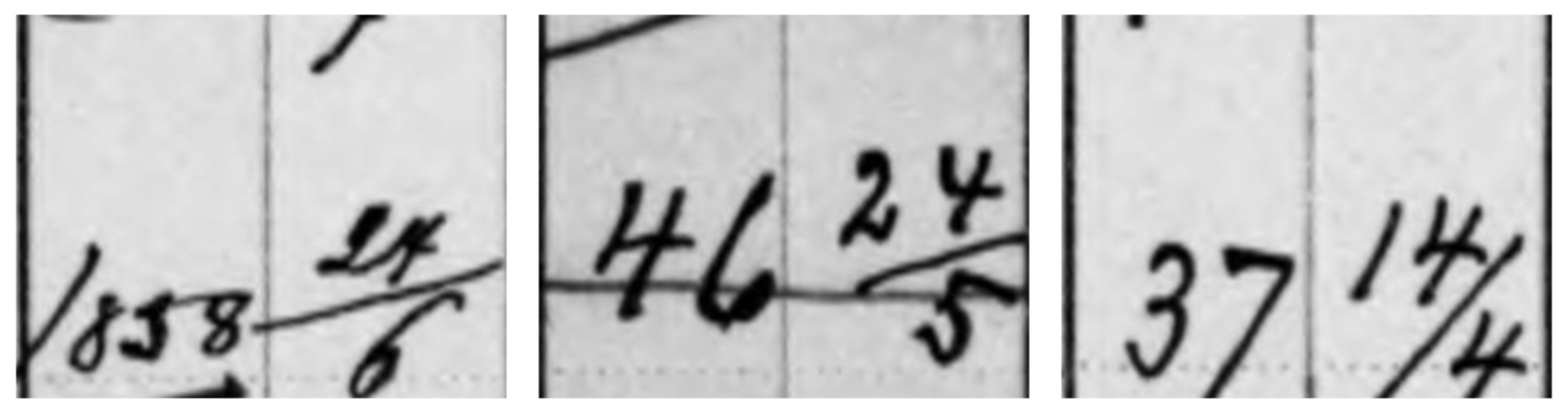}
        \label{fig:examplesSWE-BD}
    }
    \qquad
    \subfloat[Swedish Death Dates (SWE-DD)]{
        \includegraphics[width=0.4\textwidth]{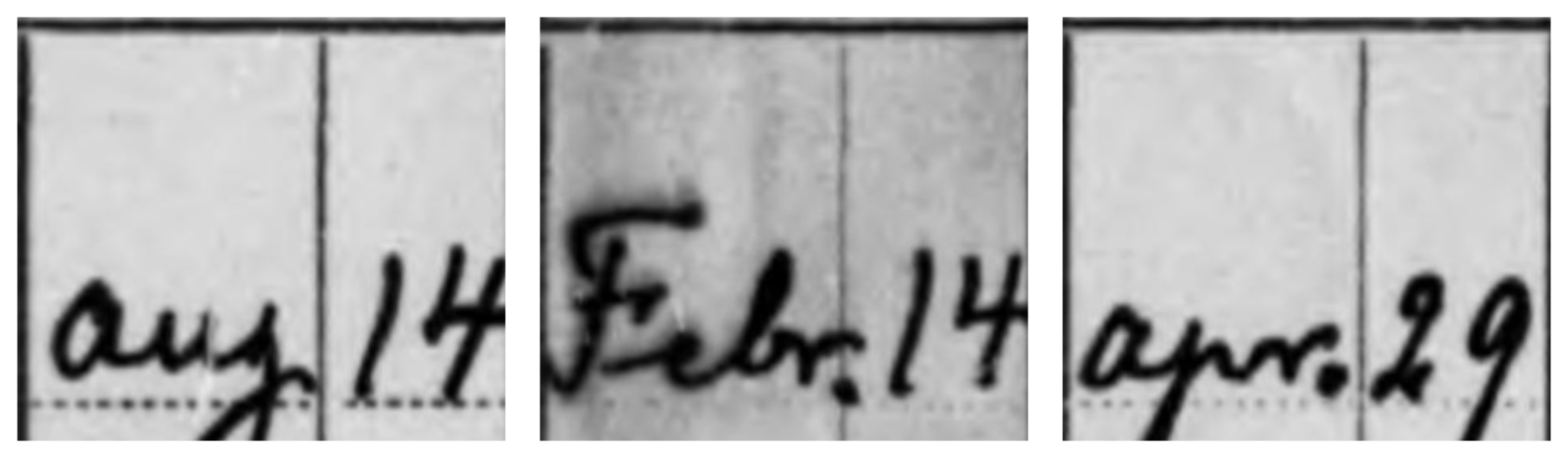}
        \label{fig:examplesSWE-DD}
    }
    \qquad
    \begin{minipage}{1\linewidth}
    \linespread{1.0}\selectfont
        \vspace{1ex}
        \footnotesize{
        \textit{Notes:}
        The figure shows a collection of examples from the DARE database, with the exception of examples from the NHVD (due to data confidentially agreements). 
        Panel \ref{fig:examplesDC-1} show examples from the DC-1 dataset. 
        Panel \ref{fig:examplesDC-2} shows examples from the DC-2 dataset.  
        Panel \ref{fig:examplesPR-1} shows examples from the PR-1 dataset. 
        Panel \ref{fig:examplesPR-2} and \ref{fig:examplesPR-2-1} show examples from the PR-2 dataset. 
        Panel \ref{fig:examplesFR} shows examples from the FR-1 dataset. 
        Panel \ref{fig:examplesSWE-BD} shows examples from the SWE-BD dataset. 
        Panel \ref{fig:examplesSWE-DD} shows examples from the SWE-DD dataset.
        }
    \end{minipage}
\end{figure*}

The DARE database consists of eight datasets.
We use Danish death certificates, Danish police records, Danish funeral records, Danish nurse home visit documents, and Swedish causes of death records. Figure \ref{fig: database-image-examples} provides visual representations of each data source.
From these sources, we define eight datasets, as the Danish death certificates, the Danish police records, and the Swedish causes of death each consist of two separate datasets.
The first dataset from the Danish death certificates contains images labelled at the University of Southern Denmark and the second contains images labelled at the Danish National Archives.
The police records are also split into two datasets, where the first refers to the date of birth of the primary person on a sheet and the second refers to the date of birth of any other person registered on the sheet (i.e., the spouse or children).
For the Swedish registers the first dataset refers to the day of birth of the person and the second to the day of death.

\subsection{Image Segmentation}
To segment the data (i.e., extract the parts of the scanned pages that contain dates), we use point set registration. 
Point set registration refers to the problem of aligning point spaces across a template image to an input image \citep{registration}.
To find point spaces that roughly correspond to each other across semi-structured documents, we extract horizontal and vertical lines from the documents.

We use the intersections as the point space, which we align with the template points. 
This method is identical to the one described in \citet{dahl2021applications}.

\subsection{Out-of-Distribution: Swedish Grade Sheets}
To illustrate how the DARE system can be used for transfer learning, we use data from Swedish grade sheets digitized as part of the ATLASS project \citep{dahl2022atlass}.
From ATLASS, we obtain 6,139 images that contain dates of the formats DD-M, DD-M-YY, DD-M-YYYY, and YYYY, with most images containing dates with the format DD-M-YY.
For our transfer learning experiments, we use 2,000 of these images as training data and the remaining 4,139 as a test set.\footnote{Here, the objective is to show the performance in a setting where only few labelled images are available, which motivated the choice of the small training sample. Further, to properly show the increase in performance when using transfer learning, we want a sufficiently large test sample.}

\subsection{Out-of-Distribution: Danish Census Data}
\begin{figure}
    \captionsetup{width=.8\textwidth}
    \caption{Example of Danish Census Data}
    \label{fig:census}
    \centering
    \includegraphics[width=0.6 \textwidth]{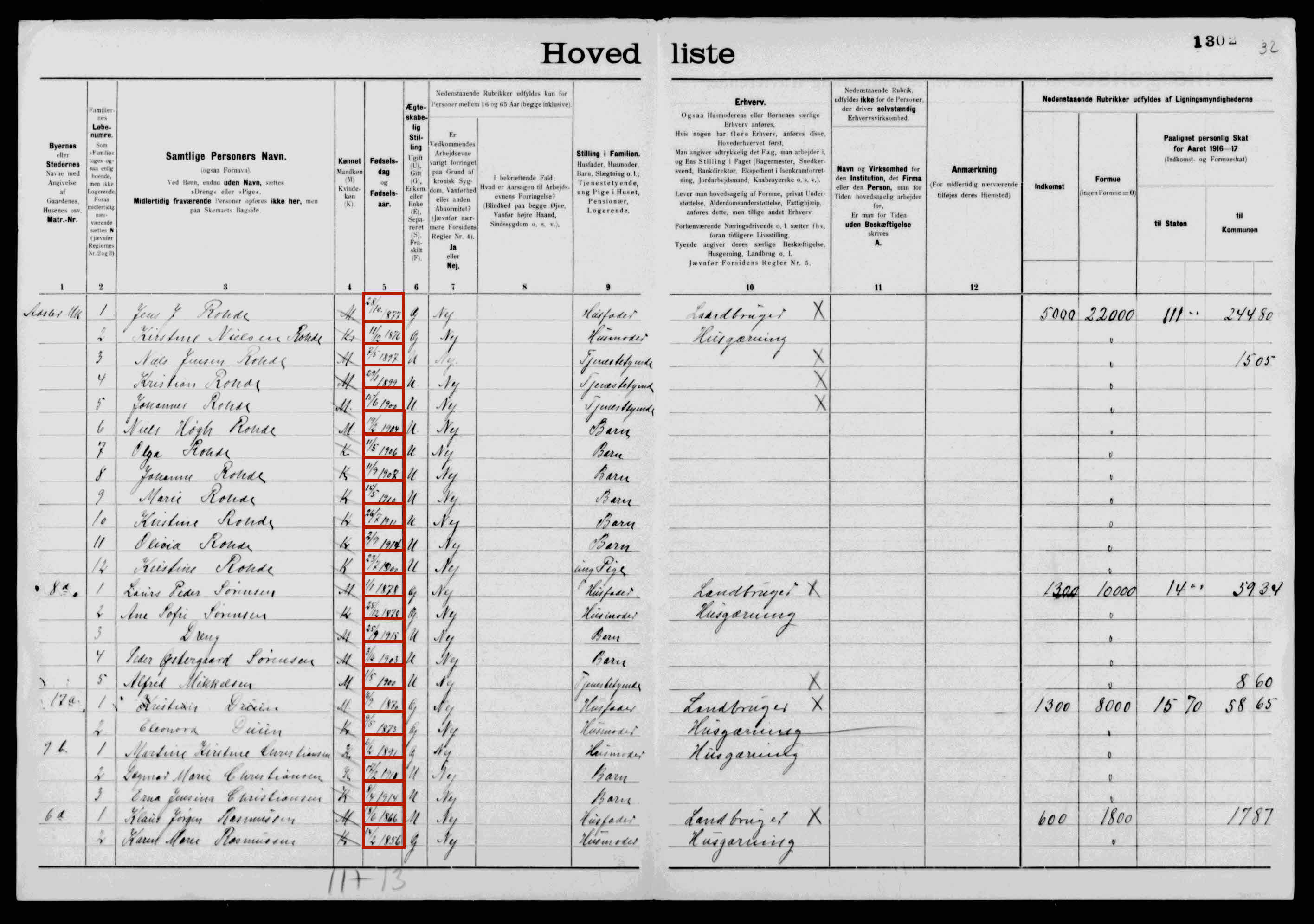}
    \begin{minipage}{1\linewidth}
        \vspace{1ex}
        \scriptsize{
        \textit{Notes:}
        The figure shows an example of a census document from the 1916 Danish Census provided by the National Archives of Denmark. This specific document contains information on 24 separate individuals from five separate families with birth dates between 1856 and 1915 (the dates are highlighted). The document appears similar in structure to the Swedish cause of death register form).
        %forms (see Figure \ref{fig:swedish}).
        }
    \end{minipage}
\end{figure}

%The Danish census data dates back to 1787 and was usually registered every five or ten years. 
We use data only from the 1916 Danish Census (excluding Copenhagen) to illustrate the performance of the DARE system for matching and linking purposes.
Figure~\ref{fig:census} shows an example of the Danish census data.
Aside from illustrating the performance of the DARE system, this dataset is also interesting in itself.
In general, census records contain information about civil status, taxation and property, birthplace, birth date, family relationships, and more. 
Due to the time period of this census, it might be of particular interest for Spanish flu research.
Further, it might help shed light on registered income, taxation, and property questions from the early 1900s in Denmark \citep{census}. 

\FloatBarrier
\newpage
\section{Model Parameters}
\label{model:params}

This section introduces the parameters of the DARE system. We use two different parameter settings with a varying number of epochs. Since there is a large difference in the sizes of the different datasets, some required more epochs for training. Model setting (2) is used in most of our applications, including the larger M-DDMYY model which is used in our transfer learning experiments.

\begin{table}[!ht]
    \centering
    \caption{Model Parameters}
    \label{tab:params}
        \footnotesize{
        \begin{tabular}{l rrrrr}
            \toprule
            \textbf{Model(s)} & \textbf{(1)} & \textbf{(2)} \\
            \midrule

RandAugment            &  $N=2,M=7$  & $N=2,M=7$ \\
Batch size             &  256 & 308 \\
Gradient clip value    &       0.02. & 0.02  \\
Dropout prob.          &        0.4 & 0.4\\
Stochastic depth prob. &       0.25 & 0.25 \\
Epochs                 & 240 & \{60, 90, 120, 240\} \\
Learning rate          & 0.5 & 0.6 \\
Momentum               & 0.9 & 0.9 \\
Random erase prob.     & 0.4 & 0.4 \\
Label smoothing        & 0.1 & 0.1 \\
Weight decay           &  0.000007 & 0.000007 \\
\bottomrule
        \end{tabular}
    }
    \begin{minipage}{1\linewidth}
            \linespread{1.0}\selectfont
        \vspace{1ex}
        \scriptsize{
            \textit{Notes:}
%            \co{Columns: (1) CIHVR, DC-1, DC-2; (2) FR, (3) PR-1, D-DDMM (4) PR-2, SWE, SWE (5) M-DDMMYYYY, DDMYY}
%            \co{TODO: State std of RA?}
%            \co{TODO: Split RA into two rows?}
%            \co{TODO: State any details on classification heads?}
        This table shows the parameters used for training the network. While most parameters are fixed for all models a few differences exist. \textit{Model 1} is used for M-NHVD, M-DC-1, and M-DC-2 which are all trained with a batch size of 256 and a learning rate of 0.5. The remaining models (Model 2) are all trained with a batch size of 308 and for a varying number of epochs. M-FR are trained for 240 epochs, M-PR-1 and M-DDM are trained for 90 epochs while M-PR-2, SWE-BD, SWE-DD are trained for 120 epochs. Finally, the two largest model M-DDMYY and M-DDMYYYY are trained for ``only'' 60 epochs.}
    \end{minipage}
\end{table}

%% file: main_database.bbl
\newcommand{\noop}[1]{}
\begin{thebibliography}{33}
\providecommand{\natexlab}[1]{#1}
\providecommand{\url}[1]{\texttt{#1}}
\expandafter\ifx\csname urlstyle\endcsname\relax
  \providecommand{\doi}[1]{doi: #1}\else
  \providecommand{\doi}{doi: \begingroup \urlstyle{rm}\Url}\fi

\bibitem[Aberdam et~al.(2021)Aberdam, Litman, Tsiper, Anschel, Slossberg,
  Mazor, Manmatha, and Perona]{Aberdam2021}
Aviad Aberdam, Ron Litman, Shahar Tsiper, Oron Anschel, Ron Slossberg, Shai
  Mazor, R~Manmatha, and Pietro Perona.
\newblock Sequence-to-sequence contrastive learning for text recognition.
\newblock \emph{Proceedings of the IEEE/CVF Conference on Computer Vision and
  Pattern Recognition}, pages 15302--15312, 2021.

\bibitem[Bertolami and Bunke(2008)]{Bertolami2008}
Roman Bertolami and Horst Bunke.
\newblock Hidden markov model-based ensemble methods for offline handwritten
  text line recognition.
\newblock \emph{Pattern Recognition}, 41\penalty0 (11):\penalty0 3452--3460,
  2008.

\bibitem[Besl and McKay(1992)]{registration}
Paul~J. Besl and Neil~D. McKay.
\newblock Method for registration of 3-d shapes.
\newblock \emph{Sensor fusion IV: control paradigms and data structures},
  1611:\penalty0 586--606, 1992.

\bibitem[Bouillon et~al.(2019)Bouillon, Ingold, and Liwicki]{Bouillon2019}
Manuel Bouillon, Rolf Ingold, and Marcus Liwicki.
\newblock Grayification: A meaningful grayscale conversion to improve
  handwritten historical documents analysis.
\newblock \emph{Pattern Recognition Letters}, 121:\penalty0 46--51, 2019.

\bibitem[Brock et~al.(2021)Brock, De, Smith, and Simonyan]{brock2021high}
Andrew Brock, Soham De, Samuel~L Smith, and Karen Simonyan.
\newblock High-performance large-scale image recognition without normalization.
\newblock \emph{International Conference on Machine Learning}, pages
  1059--1071, 2021.

\bibitem[Cascianelli et~al.(2022)Cascianelli, Cornia, Baraldi, and
  Cucchiara]{Cascianelli2022}
Silvia Cascianelli, Marcella Cornia, Lorenzo Baraldi, and Rita Cucchiara.
\newblock Boosting modern and historical handwritten text recognition with
  deformable convolutions.
\newblock \emph{International Journal on Document Analysis and Recognition
  (IJDAR)}, pages 1--11, 2022.

\bibitem[Cubuk et~al.(2020)Cubuk, Zoph, Shlens, and Le]{cubuk2020randaugment}
Ekin~D Cubuk, Barret Zoph, Jonathon Shlens, and Quoc~V Le.
\newblock {RandAugment}: {Practical} automated data augmentation with a reduced
  search space.
\newblock In \emph{Proceedings of the IEEE/CVF Conference on Computer Vision
  and Pattern Recognition (CVPR) Workshops}, June 2020.

\bibitem[Dahl et~al.(2021)Dahl, Johansen, S{\o}rensen, Westermann, and
  Wittrock]{dahl2021applications}
Christian~M Dahl, Torben~SD Johansen, Emil~N S{\o}rensen, Christian~E
  Westermann, and Simon~F Wittrock.
\newblock Applications of machine learning in document digitisation.
\newblock \emph{arXiv preprint arXiv:2102.03239}, 2021.

\bibitem[Dahl et~al.(2022{\natexlab{a}})Dahl, Johansen, Sørensen, and
  Wittrock]{dahl2022hana}
Christian~M. Dahl, Torben Johansen, Emil~N. Sørensen, and Simon Wittrock.
\newblock {HANA}: A {HA}ndwritten {NA}me database for offline handwritten text
  recognition, 2022{\natexlab{a}}.

\bibitem[Dahl et~al.(2022{\natexlab{b}})Dahl, Karlsson, Kjellsson, and
  Lindahl]{dahl2022atlass}
Christian~M. Dahl, Martin Karlsson, Gustav Kjellsson, and Mikeal Lindahl.
\newblock {ATLASS}.
\newblock \url{https://exkat.wiwinf.uni-due.de/}, 2022{\natexlab{b}}.
\newblock Accessed: 2022-07-07.

\bibitem[{Danish National Archive}(2021)]{census}
{Danish National Archive}.
\newblock Census.
\newblock
  \url{https://www.sa.dk/da/hjaelp-og-vejledning/rigsarkivets-online-vejledninger/folketaellinger-kom-godt-gang//},
  2021.
\newblock Accessed: 2022-07-07.

\bibitem[Dutta et~al.(2018)Dutta, Krishnan, Mathew, and Jawahar]{Dutta2018}
Kartik Dutta, Praveen Krishnan, Minesh Mathew, and C~V Jawahar.
\newblock Improving cnn-rnn hybrid networks for handwriting recognition.
\newblock \emph{2018 16th international conference on frontiers in handwriting
  recognition (ICFHR)}, pages 80--85, 2018.

\bibitem[Fogel et~al.(2020)Fogel, Averbuch-Elor, Cohen, Mazor, and
  Litman]{Fogel2020}
Sharon Fogel, Hadar Averbuch-Elor, Sarel Cohen, Shai Mazor, and Roee Litman.
\newblock Scrabblegan: Semi-supervised varying length handwritten text
  generation.
\newblock \emph{Proceedings of the IEEE/CVF conference on computer vision and
  pattern recognition}, pages 4324--4333, 2020.

\bibitem[Geetha et~al.(2021)Geetha, Thilagam, and Padmavathy]{Geetha2021}
R~Geetha, T~Thilagam, and T~Padmavathy.
\newblock Effective offline handwritten text recognition model based on a
  sequence-to-sequence approach with cnn–rnn networks.
\newblock \emph{Neural Computing and Applications}, 33\penalty0 (17):\penalty0
  10923--10934, 2021.

\bibitem[Goodfellow et~al.(2014)Goodfellow, Pouget-Abadie, Mirza, Xu,
  Warde-Farley, Ozair, Courville, and Bengio]{Goodfellow2014GAN}
Ian Goodfellow, Jean Pouget-Abadie, Mehdi Mirza, Bing Xu, David Warde-Farley,
  Sherjil Ozair, Aaron Courville, and Yoshua Bengio.
\newblock Generative adversarial nets.
\newblock \emph{Advances in neural information processing systems},
  27:\penalty0 369--376, 2014.

\bibitem[Goodfellow et~al.(2013)Goodfellow, Bulatov, Ibarz, Arnoud, and
  Shet]{goodfellow2013multi}
Ian~J Goodfellow, Yaroslav Bulatov, Julian Ibarz, Sacha Arnoud, and Vinay Shet.
\newblock Multi-digit number recognition from street view imagery using deep
  convolutional neural networks.
\newblock \emph{arXiv preprint arXiv:1312.6082}, 2013.

\bibitem[Graves and Schmidhuber(2008)]{Graves2008}
Alex Graves and Jürgen Schmidhuber.
\newblock Offline handwriting recognition with multidimensional recurrent
  neural networks.
\newblock \emph{Advances in neural information processing systems}, 21, 2008.

\bibitem[Graves et~al.(2006)Graves, Fernández, Gomez, and
  Schmidhuber]{Graves2006}
Alex Graves, Santiago Fernández, Faustino Gomez, and Jürgen Schmidhuber.
\newblock Connectionist temporal classification: labelling unsegmented sequence
  data with recurrent neural networks.
\newblock \emph{Proceedings of the 23rd international conference on Machine
  learning}, pages 369--376, 2006.

\bibitem[Graves et~al.(2008)Graves, Liwicki, Fernández, Bertolami, Bunke, and
  Schmidhuber]{Graves2008b}
Alex Graves, Marcus Liwicki, Santiago Fernández, Roman Bertolami, Horst Bunke,
  and Jürgen Schmidhuber.
\newblock A novel connectionist system for unconstrained handwriting
  recognition.
\newblock \emph{IEEE transactions on pattern analysis and machine intelligence
  31}, 32\penalty0 (5):\penalty0 855--868, 2008.

\bibitem[Huang et~al.(2016)Huang, Sun, Liu, Sedra, and
  Weinberger]{huang2016deep}
Gao Huang, Yu~Sun, Zhuang Liu, Daniel Sedra, and Kilian~Q Weinberger.
\newblock Deep networks with stochastic depth.
\newblock In \emph{European conference on computer vision}, pages 646--661.
  Springer, 2016.

\bibitem[Kang et~al.(2022)Kang, Riba, Rusi\~nol, Forn\'es, and
  Villegas]{Kang2022}
Lei Kang, Pau Riba, Mar\c{c}al Rusi\~nol, Alicia Forn\'es, and Mauricio
  Villegas.
\newblock Pay attention to what you read: Non-recurrent handwritten text-line
  recognition.
\newblock \emph{Pattern Recognition}, 129:\penalty0 108766, 2022.

\bibitem[Li et~al.(2021)Li, Lv, Cui, Lu, Florencio, Zhang, Li, and Wei]{Li2021}
Minghao Li, Tengchao Lv, Lei Cui, Yijuan Lu, Dinei Florencio, Cha Zhang,
  Zhoujun Li, and Furu Wei.
\newblock Trocr: Transformer-based optical character recognition with
  pre-trained models.
\newblock \emph{arXiv preprint arXiv:2109.10282}, 2021.

\bibitem[Puigcerver(2017)]{Puigcerver2017}
Joan Puigcerver.
\newblock Are multidimensional recurrent layers really necessary for
  handwritten text recognition?
\newblock \emph{2017 14th IAPR International Conference on Document Analysis
  and Recognition (ICDAR)}, 1:\penalty0 67--72, 2017.

\bibitem[Russakovsky et~al.(2015)Russakovsky, Deng, Su, Krause, Satheesh, Ma,
  Huang, Karpathy, Khosla, Bernstein, et~al.]{russakovsky2015imagenet}
Olga Russakovsky, Jia Deng, Hao Su, Jonathan Krause, Sanjeev Satheesh, Sean Ma,
  Zhiheng Huang, Andrej Karpathy, Aditya Khosla, Michael Bernstein, et~al.
\newblock Imagenet large scale visual recognition challenge.
\newblock \emph{International journal of computer vision}, 115\penalty0
  (3):\penalty0 211--252, 2015.

\bibitem[Srivastava et~al.(2014)Srivastava, Hinton, Krizhevsky, Sutskever, and
  Salakhutdinov]{srivastava2014dropout}
Nitish Srivastava, Geoffrey Hinton, Alex Krizhevsky, Ilya Sutskever, and Ruslan
  Salakhutdinov.
\newblock Dropout: a simple way to prevent neural networks from overfitting.
\newblock \emph{The journal of machine learning research}, 15\penalty0
  (1):\penalty0 1929--1958, 2014.

\bibitem[Szegedy et~al.(2016)Szegedy, Vanhoucke, Ioffe, Shlens, and
  Wojna]{szegedy2016rethinking}
Christian Szegedy, Vincent Vanhoucke, Sergey Ioffe, Jon Shlens, and Zbigniew
  Wojna.
\newblock Rethinking the inception architecture for computer vision.
\newblock In \emph{Proceedings of the IEEE conference on computer vision and
  pattern recognition}, pages 2818--2826, 2016.

\bibitem[Tan and Le(2021)]{tan2021efficientnetv2}
Mingxing Tan and Quoc~V Le.
\newblock Efficientnetv2: Smaller models and faster training.
\newblock \emph{arXiv preprint arXiv:2104.00298}, 2021.

\bibitem[Toselli et~al.(2004)Toselli, Juan, González, Salvador, Vidal,
  Casacuberta, Keysers, and Ney]{Toselli2004}
Alejandro~H Toselli, Alfons Juan, Jorge González, Ismael Salvador, Enrique
  Vidal, Francisco Casacuberta, Daniel Keysers, and Hermann Ney.
\newblock Integrated handwriting recognition and interpretation using
  finite-state models.
\newblock \emph{International Journal of Pattern Recognition and Artificial
  Intelligence}, 18\penalty0 (4):\penalty0 519--539, 2004.

\bibitem[Vaswani et~al.(2017)Vaswani, Shazeer, Parmar, Uszkoreit, Jones, Gomez,
  Kaiser, and Polosukhin]{Vaswani2017}
Ashish Vaswani, Noam Shazeer, Niki Parmar, Jakob Uszkoreit, Llion Jones,
  Aidan~N Gomez, Łukasz Kaiser, and Illia Polosukhin.
\newblock Attention is all you need.
\newblock \emph{Advances in neural information processing systems}, pages
  5998--6008, 2017.

\bibitem[Wightman(2021)]{rw2021timm}
Ross Wightman.
\newblock Pytorch image models.
\newblock \url{https://github.com/rwightman/pytorch-image-models}, 2021.
\newblock Accessed: 2022-07-07.

\bibitem[Wigington et~al.(2018)Wigington, Tensmeyer, Davis, Barrett, Price, and
  Cohen]{Curtis2018}
Curtis Wigington, Chris Tensmeyer, Brian Davis, William Barrett, Brian Price,
  and Scott Cohen.
\newblock Start, follow, read: End-to-end full-page handwriting recognition.
\newblock \emph{Proceedings of the European Conference on Computer Vision
  (ECCV)}, pages 367--383, 2018.

\bibitem[Yousef and Bishop(2020)]{Yousef2020}
Mohamed Yousef and Tom~E Bishop.
\newblock Origaminet: weakly-supervised, segmentation-free, one-step, full page
  text recognition by learning to unfold.
\newblock \emph{Proceedings of the IEEE/CVF Conference on Computer Vision and
  Pattern Recognition}, pages 14710--14719, 2020.

\bibitem[Zhong et~al.(2020)Zhong, Zheng, Kang, Li, and Yang]{zhong2020random}
Zhun Zhong, Liang Zheng, Guoliang Kang, Shaozi Li, and Yi~Yang.
\newblock Random erasing data augmentation.
\newblock In \emph{Proceedings of the AAAI Conference on Artificial
  Intelligence}, volume~34, pages 13001--13008, 2020.

\end{thebibliography}
